\pdfoutput=1

\documentclass[11pt]{article}

\usepackage{ACL2023}

\usepackage{times}
\usepackage{latexsym}

\usepackage[T1]{fontenc}

\usepackage[utf8]{inputenc}

\usepackage{microtype}

\usepackage{inconsolata}
\usepackage{xcolor}         %
\usepackage{tabularx}
\definecolor{bluegray}{rgb}{0.4, 0.6, 0.8}
\definecolor{cornflowerblue}{rgb}{0.39, 0.58, 0.93}
\definecolor{mayablue}{rgb}{0.21,0.49,0.74}
\definecolor{caribbeangreen}{rgb}{0.0, 0.8, 0.6}
\definecolor{candypink}{rgb}{0.89, 0.44, 0.48}

\usepackage{comment} 
\usepackage{graphicx}
\usepackage[ruled,norelsize]{algorithm2e}
\usepackage{algpseudocode}
\usepackage{algorithmicx}
\usepackage[utf8]{inputenc} %
\usepackage[T1]{fontenc}    %
\usepackage{hyperref}

\hypersetup{
    colorlinks=true,
    linkcolor=cornflowerblue,
    filecolor=magenta,      
    urlcolor=teal,
    citecolor=mayablue,
    pdftitle={A Study on the Calibration of In-context Learning},
    pdfpagemode=FullScreen,
}
   
\usepackage{subfigure}
\usepackage{url}            
\usepackage{booktabs}       
\usepackage{amsfonts}       %
\usepackage{nicefrac}       %
\usepackage{microtype}      %
\usepackage{xcolor}         %
\usepackage{graphicx}
\usepackage[toc,page,header]{appendix}
\usepackage{minitoc}
\usepackage{makecell}
\usepackage{adjustbox}
\usepackage{colortbl}
\definecolor{Gray}{gray}{0.92}
\usepackage{amsmath}
\usepackage{amssymb}
\usepackage{mathtools}
\usepackage{amsthm}
\usepackage{multirow}

\usepackage{enumitem}
\usepackage{bbm}

\usepackage[capitalize,noabbrev]{cleveref}

\theoremstyle{plain}
\usepackage{mdframed}
\newmdtheoremenv[linewidth=0pt,innerleftmargin=4pt,innerrightmargin=4pt]{prop}{Proposition}

\theoremstyle{definition}

\theoremstyle{remark}

\newcommand{\resred}[1]{\textcolor{candypink}{#1}}

\usepackage{wrapfig}

\title{A Study on the Calibration of In-context Learning}

\author{Hanlin Zhang$^{1}$\ 
  Yi-Fan Zhang$^{2}$ \
  Yaodong Yu$^{3}$ \
  \textbf{Dhruv Madeka}$^{4}$ \\
  \textbf{Dean Foster}$^{4}$ \
  \textbf{Eric Xing}$^{2,5}$ \
  \textbf{Himabindu Lakkaraju}$^{1}$ \
  \textbf{Sham Kakade}$^{1,4}$ \
  \\
  \vspace{-0.1in}
  \\
  $^1$Harvard University 
  $^2$MBZUAI
  $^3$UC Berkeley \\
  $^4$Amazon
  $^5$Carnegie Mellon University
  \\
}

\begin{document}
\maketitle
\begin{abstract}

Accurate uncertainty quantification is crucial for the safe deployment of machine learning models, and prior research has demonstrated improvements in the calibration of modern language models (LMs). 
We study in-context learning (ICL), a prevalent method for adapting static LMs through tailored prompts, and examine the balance between performance and calibration across a broad spectrum of natural language understanding and reasoning tasks. 
Through comprehensive experiments, we observe that, with an increasing number of ICL examples, models initially exhibit increased miscalibration before achieving better calibration and miscalibration tends to arise in low-shot settings.
Moreover, we find that methods aimed at improving usability, such as fine-tuning and chain-of-thought (CoT) prompting, can lead to miscalibration and unreliable natural language explanations. 
Furthermore, we explore recalibration techniques and find that a scaling-binning calibrator can reduce calibration errors consistently.

\end{abstract}

\section{Introduction}\label{sec:intro}

Language models (LMs) that encompass transformer-based architectures \citep{brown2020language, chowdhery2023palm, openai2023gpt4} can generate coherent and contextually relevant texts for various use cases.
Despite their impressive performance, these models occasionally produce erroneous or overconfident outputs, leading to concerns about their calibration \citep{dawid1982well, degroot1983comparison} which measures how faithful a model's prediction uncertainty is.
Such a problem is pressing when users adapt them using a recent paradigm called in-context learning \citep{brown2020language} to construct performant predictors, especially for applications in safety-critical domains \citep{bhatt2021uncertainty, pan2023rewards}.

We provide an in-depth evaluation and analysis of how well these models are calibrated - that is, the alignment between the model's confidence in its predictions and the actual correctness of those predictions. 
This token-level calibration assessment enables us to measure the discrepancy between the model's perceived and actual performance to assess its accuracy and reliability through a Bayesian uncertainty lens. 
\begin{figure*}[h]
\centering
\begin{minipage}[t]{\textwidth}
\subfigure[Demonstration of In-context Learning]{
\includegraphics[width=0.33\textwidth]{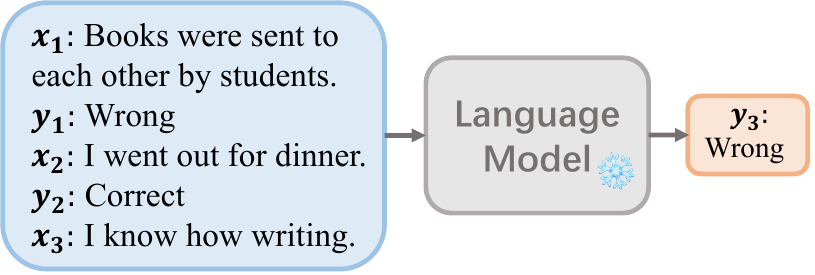} 
\label{fig:icl}
} 
\subfigure[The accuracy and calibration of LLaMA-7B]{
\includegraphics[width=.32\textwidth]{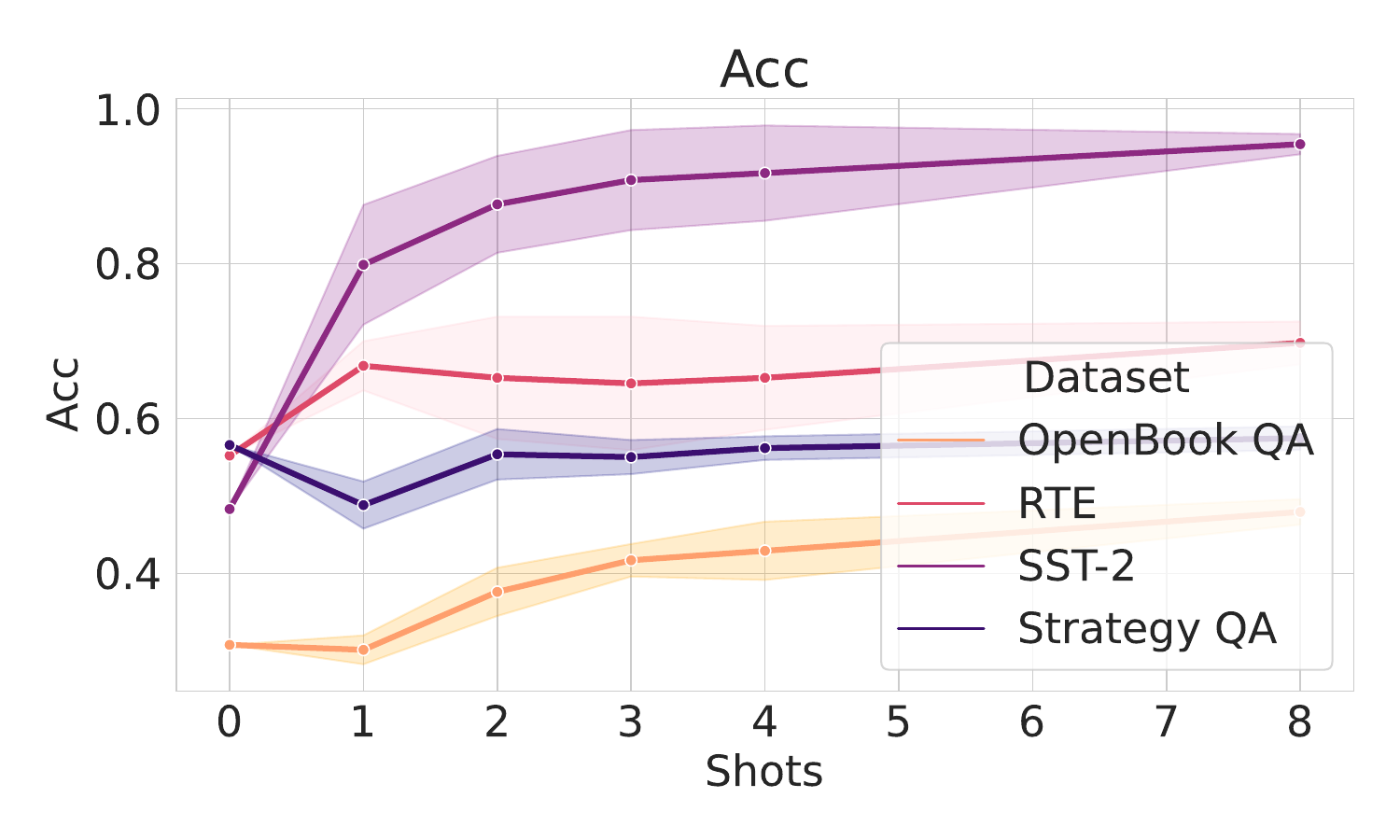} %
\includegraphics[width=.32\textwidth]{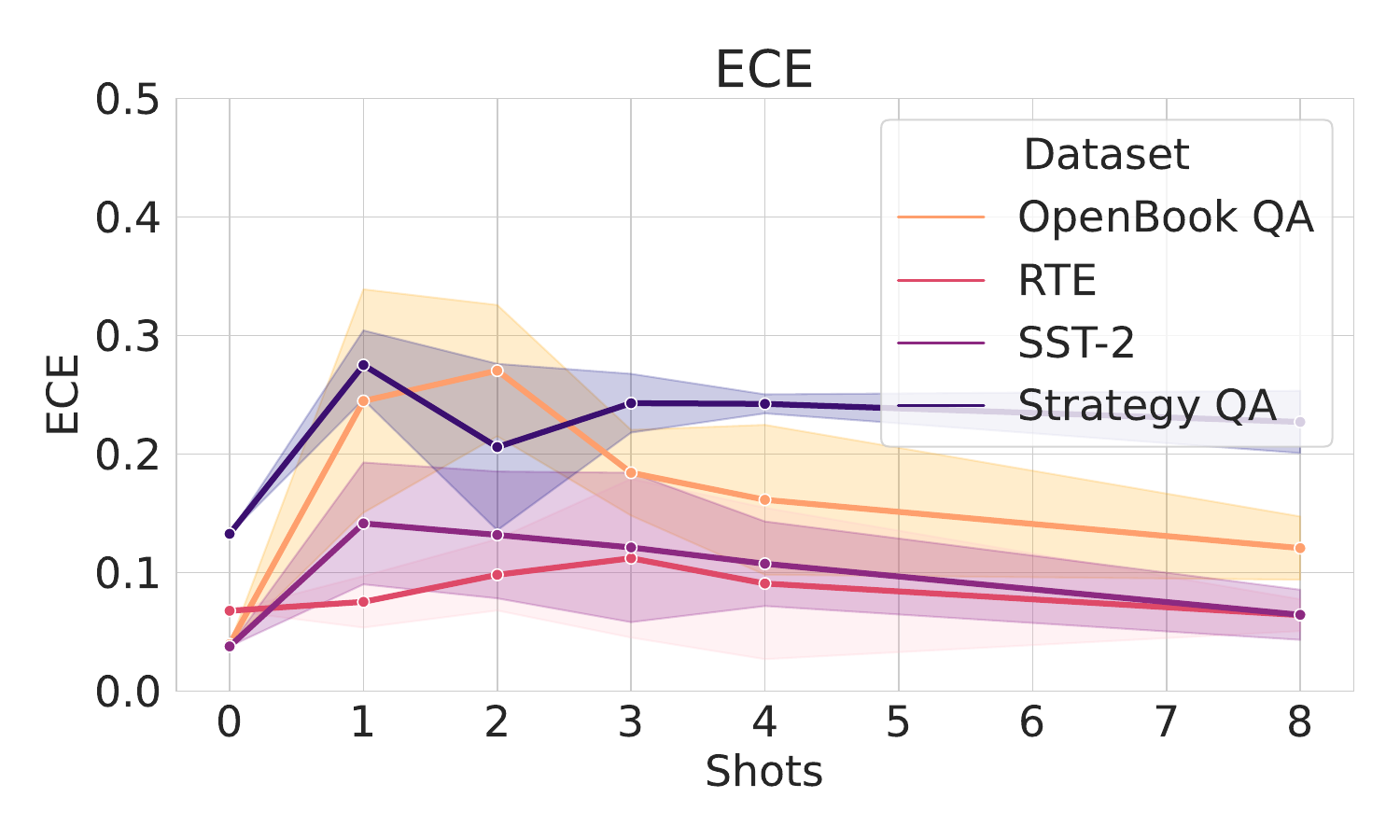}
}
\end{minipage}
\caption{ \textbf{The accuracy-calibration trade-off of in-context learning.} 
(a) ICL concerns taking task-specific examples as the prompt to adapt a frozen LLM to predict the answer.
(b) Classification accuracy and expected calibration error of ICL. 
As the number of ICL samples increases, the prediction accuracy improves (\textbf{Left}); at the same time, the calibration first worsens ($k<3$) and then becomes better (\textbf{Right}). 
}
\label{fig:teaser}
\end{figure*}

We find that LM such as LLaMA \citep{touvron2023llama} is poorly calibrated in performant settings and there exists a calibration-accuracy trade-off (Fig.\ref{fig:teaser}) for low-shot settings ($k<4$): as we increase the amount of in-context samples, both prediction accuracy and calibration error increase. Such a trade-off can be improved using more ICL examples ($k=8$) and larger models.
Crucially, this calibration degradation worsens when fine-tuning occurs using specialized data to improve usability, such as curated instructions \citep{dubois2023alpacafarm}, dialogues \citep{zheng2023judging}, or human preference data \citep{ziegler2019fine}.
Though previous common practice suggests recalibrating models' logits via temperature scaling \citep{guo2017calibration}, we show that in contrast to classic regimes, the miscalibration issue in ICL can not be easily addressed using such well-established scaling approaches \citep{platt1999probabilistic}. 
Thus we propose to use scaling-binning \citep{kumar2019verified}, which fits a scaling function, bins its outputs, and then outputs the average of the function values in that bin, to reduce the expected calibration error below $0.1$.

Furthermore, we study the trade-off in reasoning tasks that involve generation of explanations \citep{camburu2018snli, nye2021show, wei2022chain} before the answer, showing that the model can produce confidently wrong answers (using confidence histograms and reliability plots) when prompted with explanations on Strategy QA \citep{geva2021did}, Commonsense QA \citep{talmor2018commonsenseqa}, OpenBook QA \citep{OpenBookQA2018}, World Tree \citep{jansen2018worldtree}. 
We carefully design our human evaluation and observe that, with the increase in model sizes and the quantity of ICL examples, there is a corresponding rise in the proportion of confidently predicted examples among those incorrectly forecasted. Moreover, we find that a high proportion of wrong predictions are of high confidence and showcase those typical confidently wrong examples of LMs.

Moreover, we find that choosing ICL samples from the validation set does not naturally lead to calibrated predictions, showing that ICL learns in a fairly different way than stochastic gradient descent, a common prototype previous works hypothesize \citep{von2023transformers}.
Motivated by this difficulty, we design controlled experiments to illustrate that when examples in the prompt are sampled from the same task instead of repeating a given example in various ways, the learning performance would be improved.

\section{Related Work}\label{sec:related}

\textbf{Calibration of language models.}
Calibration is a safety property to measure the faithfulness of machine learning models' uncertainty, especially for error-prone tasks using LMs.
Previous works find that pre-training \citep{desai2020calibration} and explanation \citep{zhang2020effect, gonzalez2021explanations} improves calibration.
Models can be very poorly calibrated when we prompt LMs \citep{jiang2021can}, while calibration can also depend on model size \citep{kadavath2022language}.
\citep{braverman2020calibration} assesses the long-term dependencies in a language model’s generations compared to those of the underlying language and finds that entropy drifts as models such as when GPT-2 generates text.
The intricacy of explanations on complementary team performance poses additional challenges due to the overreliance on explanations of users regardless of their correctness \citep{bansal2021does}.
\citep{mielke2022reducing} gives a framework for \textit{linguistic calibration}, a concept that emphasizes the alignment of a model's expressed confidence or doubt with the actual accuracy of its responses. 
The process involves annotating generations with \texttt{<DK>}, \texttt{<LO>}, \texttt{<HI>} for confidence levels, then training the confidence-controlled model by appending the control token \texttt{<DK/LO/HI>} at the start of the output, followed by training a calibrator to predict these confidence levels, and finally predicting confidence when generating new examples.
\citep{tian2023just} finds that asking LMs for their probabilities can be better than using conditional probabilities in a traditional way. 
LHTS \citep{shih2023long} is a simple amortized inference trick for temperature-scaled sampling from LMs and diffusion models. 
To aggregate log probabilities across semantically equivalent outputs, \citet{kuhn2023semantic} utilize bidirectional entailment through a model to identify outputs that are semantically similar, thereby refining the uncertainty estimation process.
\citep{cole2023selectively} identifies the calibration challenge in ambiguous QA and distinguishes uncertainty about the answer (epistemic uncertainty) from uncertainty about the meaning of the question (denotational uncertainty), proposing sampling and self-verification methods.
\citet{kamath2020selective} trains a calibrator to identify inputs on which the QA model errs and abstains when it predicts an error is likely.
\citet{zhao2023automatic} proposes the Pareto optimal learning assessed risk score for calibration and error correction but requires additional training.
\citet{kalai2023calibrated} show the trade-off between calibration and hallucination but they didn't study it in a realistic setting and how the predicted answer's accuracy would impact those two safety aspects.

\noindent
\textbf{In-context learning.}
Large models such as GPT-3 \citep{brown2020language} have demonstrated the potential of in-context learning, a method where the model infers the task at hand from the context provided in the input, without requiring explicit retraining or fine-tuning for each new task.
Some recent works attempt to understand ICL through meta-learning \citep{von2023transformers}, Bayesian inference \citep{xie2021explanation}, mechanistic interpretability \citep{olsson2022context},  algorithm selection \citep{bai2023transformers}, synthetic data and simple function classes \citep{garg2022can, akyurek2022learning, raventos2023pretraining}.
Notably, unlike previous works \citep{zhao2021calibrate, han2023prototypical, fei2023mitigating, zhou2023batch} that focus on improving task accuracy using the same ``\textit{calibration}'' terminology, we study the uncertainty of ICL and measure its trade-off with accuracy.

\section{Background}\label{sec:background}

\begin{table*}[h]
\centering
\resizebox{\textwidth}{!}{%
\begin{tabular}{c|c|c|c|c|c|c|c|c|c|c|c|c}
\toprule \hline
\multirow{3}{*}{Dataset} & \multicolumn{12}{c}{LLaMA-30B} \\
\cline{2-13}
& \multicolumn{2}{c|}{0-shot} & \multicolumn{2}{c|}{1-shot} & \multicolumn{2}{c|}{2-shot} & \multicolumn{2}{c|}{3-shot} & \multicolumn{2}{c|}{4-shot} & \multicolumn{2}{c}{8-shot} \\
\cline{2-13}
& ECE & Acc & ECE & Acc & ECE & Acc & ECE & Acc & ECE & Acc& ECE & Acc  \\
\hline
& \multicolumn{12}{c}{Text Classification} \\
\hline
AGNews & 0.261 & 0.37 & 0.043 & 0.830 & 0.049 & 0.817 & \resred{0.067} & 0.810 & 0.049 & 0.821 & 0.047 & 0.855 \\
\hline
RTE & 0.023 & 0.672 & 0.051 & 0.742 & \resred{0.060} & 0.747 & 0.050 & 0.738 & 0.048 & 0.748 & 0.058 & 0.752 \\
\hline
CB & 0.069 & 0.500 & \resred{0.312} & 0.696 & 0.216 & 0.789 & 0.217 & 0.834 & 0.192 & 0.814 & 0.181 & 0.796 \\
\hline
SST-2 & 0.083 & 0.607 & \resred{0.163} & 0.930 & 0.139 & 0.940 & 0.126 & 0.961 & 0.112 & 0.964 & 0.080 & 0.964 \\
\hline
& \multicolumn{12}{c}{Reasoning with Scratchpad} \\
\hline
Strategy QA & 0.204 & 0.450 & 0.154 & 0.619 & 0.174 & 0.654 & \resred{0.172} & 0.660 & 0.161 & 0.672 & 0.152  & 0.665  \\
 \hline
Commonsense QA & 0.048 & 0.356 & 0.232 & 0.589 & \resred{0.290} & 0.608 & 0.253 & 0.675 & 0.283 & 0.644 & 0.289 &  0.653 \\
 \hline
World Tree & 0.112 & 0.534 & 0.211 & 0.570 & \resred{0.251} & 0.621 & 0.185 & 0.680 & 0.206 & 0.646 & - & - \\ %
 \hline
OpenBook QA & 0.036 & 0.386 & 0.231 & 0.561 & \resred{0.255} & 0.604 & 0.207 & 0.644 & 0.206 & 0.648 & 0.191 & 0.662 \\
 \hline
\bottomrule 
\end{tabular}
}
\caption{\textbf{Accuracy and Calibration} of LLaMA-30B model with across four text classification datasets and four reasoning datasets. Results are excluded when the data exceeds the context length limit.}
\label{tab:main_tab}
\end{table*}

\begin{figure*}[thb]
\begin{minipage}[t]{\textwidth}
\centering
\subfigure{
\includegraphics[width=.3\textwidth]{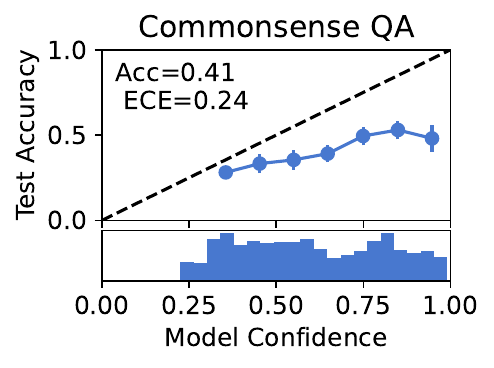}
\label{fig:gap-ece-7b}
}
\subfigure{
\includegraphics[width=.3\textwidth]{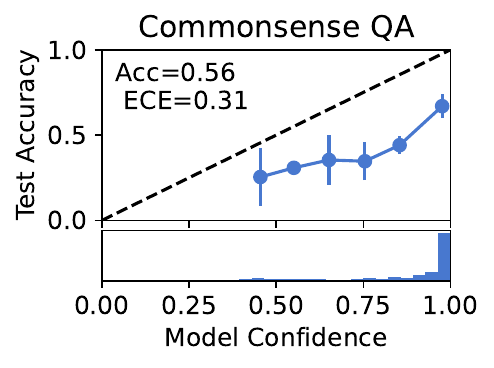} 
\label{fig:gap-loss-13b} 
}
\subfigure{
\includegraphics[width=.3\textwidth]{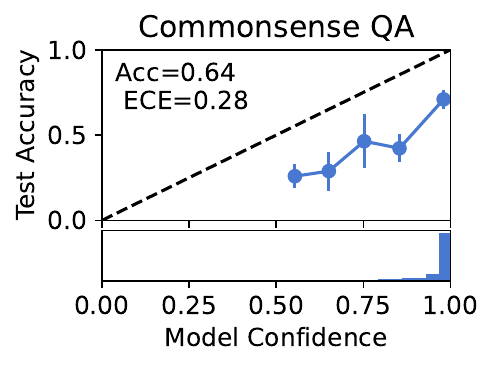} 
\label{fig:gap-loss-30b} 
}
\end{minipage}
\begin{minipage}[t]{\textwidth}
\centering
\subfigure{
\includegraphics[width=.3\textwidth]{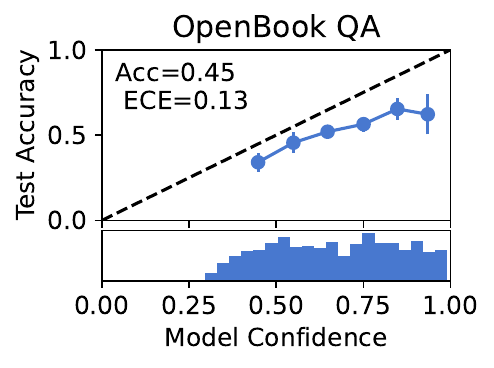}
\label{fig:gap-ece-7b}
}
\subfigure{
\includegraphics[width=.3\textwidth]{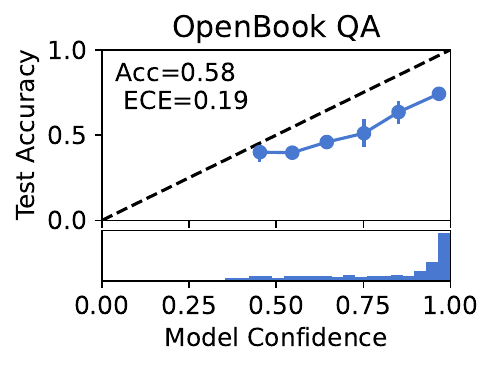} 
\label{fig:gap-loss-13b} 
}
\subfigure{
\includegraphics[width=.3\textwidth]{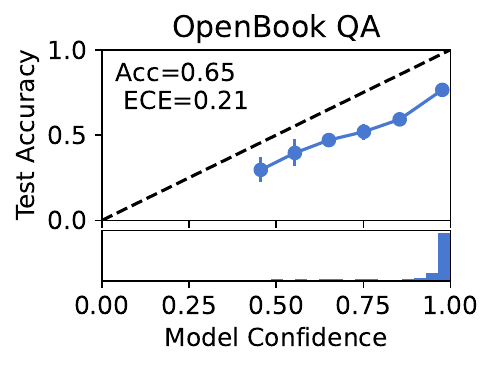} 
\label{fig:gap-loss-30b} 
}
\end{minipage}
\caption{
\textbf{Reliability plots and confidence histograms} of LLaMA models on 4-shot learning tasks. 
Results of different sizes 7B (left), 13B (middle), and 30B (right) are plotted.
}
\label{fig:reliability}
\end{figure*}

\textbf{Setting.} Given a pre-trained language model $\mathcal{P}_\theta(w_t|w_{<t})$, we seek to adapt it using the prompt $w_0 = [x_1, y_1, x_2, y_2, \dots, x_{n-1}, y_{n-1}, x_n]$ to generate a predicted answer $y_n=\mathcal{P}_\theta(w_0)$. 
In the context of reasoning, a popular approach is to hand-craft some explanations/rationales/chain-of-thoughts $e$ in the prompt $w_0 = [x_1, e_1, y_1, x_2, e_2, y_2, \dots, x_{n-1}, e_{n-1}, y_{n-1}, x_n]$ to generate explanation $e_n$ and answer $y_n$, for the test sample: $\overbrace{w_1, w_2, \dots, w_{k}}^{e_{n}}, y_n = \mathcal{P}_\theta(w_0)$.

We extract answer token probabilities of LMs, e.g. for binary classification tasks, we filter and extract probabilities $P(``\textrm{Yes}")$ and $P(``\textrm{No}")$, based on which we calculate the following statistics for studying the confidence and calibration of LMs:

\noindent
\textbf{Confidence and feature norm.} We record the maximum probability of the answer token as its confidence $\textrm{Conf}=\mathcal{P}_\theta(y_n|w_{<n})$ and the feature norm $z_n$ as the intermediate hidden state before the linear prediction layer.

\noindent
\textbf{Entropy rate.} We denote the entropy of a token $w_t$ at position $t$ as $H(w_t|w_{<t}) = - \mathbb{E}_{w_{t}\sim\mathcal{P}_{\theta}(\cdot|w_{<t})}[\log \mathcal{P}_\theta(w_t|w_{<t})]$. We typically measure it based on the answer token via setting $w_t=y_n$.
Note that auto-regressive LMs are trained via maximizing the negative log-likelihood objective $\mathcal{L}=-\mathbb{E}_{t}[\log \mathcal{P}_\theta\left(w_t|w_{<t}\right)]$ on massive corpora.

\noindent
\textbf{Empirical estimate of the expected calibration error (ECE)} In the realm of probabilistic classifiers, calibration is a crucial concept. A classifier, denoted as $\mathcal{P}_\theta$ with parameters $\theta$ and operating over $C$ classes, is said to be "canonically calibrated"~\citep{kull2015novel} when, for every probability distribution $p$ over the $C$ classes and for every label $y$, the probability that the label is $y$ given the classifier's prediction is $p$ matches the component of $p$ corresponding to $y$. This is mathematically represented as, $\forall p \in \Delta^{C-1}, \forall y \in Y:$
\begin{equation}
\resizebox{.3\textwidth}{!}{
$P\left(Y=y \mid \mathcal{P}_\theta(X)=p\right)=p_y.$
}
\end{equation}
Here, $\Delta^{C-1}$ symbolizes the $(C-1)$-dimensional simplex, which encompasses all potential probability distributions over the $C$ classes.

A simpler calibration criterion is the "confidence calibration." In this case, a classifier is deemed calibrated if, for every top predicted probability $p^*$, the probability that the true label belongs to the class with the highest predicted probability, given that this maximum predicted probability is $p^*$, equals $p^*$. Formally:  $\forall p^* \in[0,1], $
\begin{equation}
\resizebox{.4\textwidth}{!}{
$
P\left(Y = c(X) \mid \max \mathcal{P}_\theta(X)=p^*\right)=p^*,$
}
\label{eq:def}
\end{equation}
where $c(X) = \arg\max p$ and  ties are
broken arbitrarily.  
To gauge the calibration of a model, we adopt Expected Calibration Error (ECE~\cite{guo2017calibration}) defined as:

\begin{equation}
\resizebox{.42\textwidth}{!}{
$\mathbb{E}\left[\left|p^*-\mathbb{E}\left[Y =c(X) \mid \max \mathcal{P}_\theta(X)=p^*\right]\right|\right].$
}
\end{equation}

In real-world applications, this quantity cannot be computed without quantization. So, the ECE is approximated by segmenting predicted confidences into $M$ distinct bins, $B_1, \ldots, B_M$. The approximation is then computed as:
$$
\widehat{\mathrm{ECE}}=\sum_{m=1}^M \frac{\left|B_m\right|}{n}\left|\operatorname{acc}\left(B_m\right)-\operatorname{conf}\left(B_m\right)\right|.
$$
Here, $\operatorname{acc}\left(B_m\right)$ is the accuracy within bin $B_m$, and $\operatorname{conf}\left(B_m\right)$ is the average confidence of predictions in bin $B_m$. The total number of samples is represented by $n$, and the dataset consists of $n$ independent and identically distributed samples, $\left\{\left(x_i, y_i\right)\right\}_{i=1}^n$. In our work, we use this estimator to approximate the ECE.

\section{Experiments}\label{sec:exp}

We briefly summarize our results and findings before explaining the experimental settings.
\begin{itemize}[leftmargin=*]
    \item For the base LMs we considered, they are calibrated when prompting with a sufficient amount of ICL examples to get non-trivial performance. 
    \item As we increase the number of ICL examples, models tend to be first more miscalibrated and then calibrated. In low-shot settings ($k<4$), models can be mis-calibrated, in part due to poor data (aleatoric) uncertainty. 
    \item Interventions that improve usability such as fine-tuning, and chain-of-thought (CoT) prompting would lead to miscalibration. The generated explanations from CoT can improve predictive results but may not be reliable by human evaluation.
\end{itemize}

\subsection{Experimental Settings}

\textbf{Models.} We study decoder-only autoregressive LMs involving LLaMA \citep{touvron2023llama}, ranging from 7B to 30B, and its variants fine-tuned with instruction, dialog, or RLHF like Alpaca \citep{dubois2023alpacafarm}, Vicuna \citep{zheng2023judging}, and LLaMA2-Chat \citep{touvron2023llama2}.

\noindent
\textbf{Datasets and tasks.}  %
We used both traditional NLU tasks such as AGNews \citep{zhang2015character}, TREC \citep{voorhees2000building}, CB \citep{schick2021exploiting}, SST-2 \citep{socher2013recursive}, DBPedia \citep{zhang2015character}, 
as well as reasoning question answering tasks like Strategy QA \citep{geva2021did}, Commonsense QA \citep{talmor2018commonsenseqa}, OpenBook QA \citep{OpenBookQA2018}, World Tree \citep{jansen2018worldtree}. 
Notably, the reasoning task performance can be greatly improved in general via prompting methods like scratchpad \citep{nye2021show, wei2022chain} that enables models to generate natural language explanations before predicting an answer.

\noindent
\textbf{In-context learning settings.}
For $k$-shot learning, we prompt the model via sampling $k$ examples from the training set for each test example. 
Each experiment is repeated 10 times to reduce variance and we report the mean results. 
We use $M=10$ bins for calculating calibration errors.

\subsection{Numerical Results}

\textbf{Model performance and calibration.}  We record the performance and calibration errors for $k$-shot learning ($k=0,1,2,3,4,8$), characterizing the calibration-accuracy trade-off in both classic and realistic settings (Tab.~\ref{tab:main_tab}). 
Our findings are two-fold: as more in-context examples are included, we observe a concurrent rise in both accuracy and calibration error across most low-shot situations. Especially, when $k=0$ increases to $k=1$, there is a marked boost in both accuracy and calibration error, demonstrating the importance of in-context examples in learning performance while one single example may not be able to reduce aleatoric uncertainty. 
In particular, for reasoning tasks, we explore prompting approaches that explicitly include explanations in reasoning tasks, i.e. scratchpad \citep{nye2021show} or chain-of-thought \citep{wei2022chain}, showing that calibration significantly degrades after generating a long context for reasoning and explaining the final answer. 
We also note that having more ICL examples does not necessarily lead to better calibration though the predictive performance can generally improve (e.g., $k=8$ for CB in Tab.\ref{tab:main_tab}).
This may stem from the intrinsic limitations of transformers in effectively modeling long-term dependencies.

\begin{table*}[]
\centering
\resizebox{0.9\textwidth}{!}{%
\begin{tabular}{ccccccccccc}
\toprule \hline
 & \multicolumn{2}{c}{1-shot} & \multicolumn{2}{c}{2-shot} & \multicolumn{2}{c}{4-shot} & \multicolumn{2}{c}{8-shot} & \multirow{2}{*}{Avg Acc} & \multirow{2}{*}{Avg ECE} \\ \cmidrule(lr){2-3} \cmidrule(lr){4-5}  \cmidrule(lr){6-7}   \cmidrule(lr){8-9} 
 & ACC & ECE & ACC & ECE & ACC & ECE & ACC & ECE &  &  \\ \hline
Vanilla & \textbf{0.740} & \textbf{0.098} & \textbf{0.877} & \textbf{0.132} & \textbf{0.917} & \textbf{0.108} & \textbf{0.954} & \textbf{0.064} & \textbf{0.872} & \textbf{0.100} \\
Repeat prompt & 0.740 & 0.098 & 0.693 & 0.155 & 0.801 & 0.117 & 0.820 & 0.111 & 0.764 & 0.120 \\
Repeat context & 0.740 & 0.098 & 0.668 & 0.208 & 0.657 & 0.220 & 0.607 & 0.219 & 0.668 & 0.186 \\ \hline
\bottomrule
\end{tabular}%
}
\caption{\textbf{Acc} and \textbf{ECE} of LLaMA-7B model on SST-2 with different prompt repetition strategies.}\label{tab:prompt_rep}
\end{table*}

\noindent
\textbf{Post-hoc recalibraiton.}
We conducted experiments with three strategies (Algorithm~\ref{alg:temperature_scaling}) to address miscalibration using temperature scaling~\citep{guo2017calibration} and scaling-binning~\citep{kumar2019verified} with learnable parameter $w$: 
\begin{enumerate}
\item (\textbf{$0$-shot}) Learning $w$ from the training split and applying it to all test samples with different shot numbers.
\item (\textbf{$k$-shot}) Learning $w$ for each $k$-shot ICL; in other words, different temperatures are learned for different shot numbers in ICL.
\item (\textbf{Fix $w$}) Fixing the prompt for each experiment and learning $w$ corresponding to the fixed prompt. In other words, $w$ is learned for calibration for every possible ICL prompt.
\end{enumerate}
In Appendix Alg. \ref{alg:temperature_scaling}, we introduce the recalibration algorithm employing temperature scaling. Additionally, we utilize the scaling-binning calibrator \citep{kumar2019verified}, which fits a calibration function $w \in \mathcal{W}$ to the recalibration dataset: $\arg\min_w  \sum_{(x_i, y_i)} \ell(w\cdot\mathcal{P}_\theta(x_i), y_i)$, where $\ell$ is log-loss. Subsequently, the input space is partitioned into bins, ensuring an equal number of inputs in each bin (defaulting to 10 bins). Within each bin, the average of the $w$ values is computed and outputted for recalibration.

Upon examination of \cref{tab:temp} and \cref{tab:scal_bin}, it is evident that none of the aforementioned strategies utilizing temperature scaling achieves satisfactory calibration performance. This finding contrasts with the well-established success of scaling confidence scores in the supervised learning setting, where it effectively reduces calibration errors \citep{guo2017calibration}. The fact that applying a post-processing calibration method, such as temperature scaling, cannot directly resolve the miscalibration issue suggests that ICL might have different properties compared to predictions from classical supervised learning models.
On the other hand, the scaling-binning method demonstrates superior performance in our experiments, which successfully reduces calibration errors below $0.1$.

\begin{figure*}[h]
\centering
\begin{minipage}[t]{\textwidth}
\subfigure[Classification accuracy]{
\includegraphics[width=0.48\textwidth]{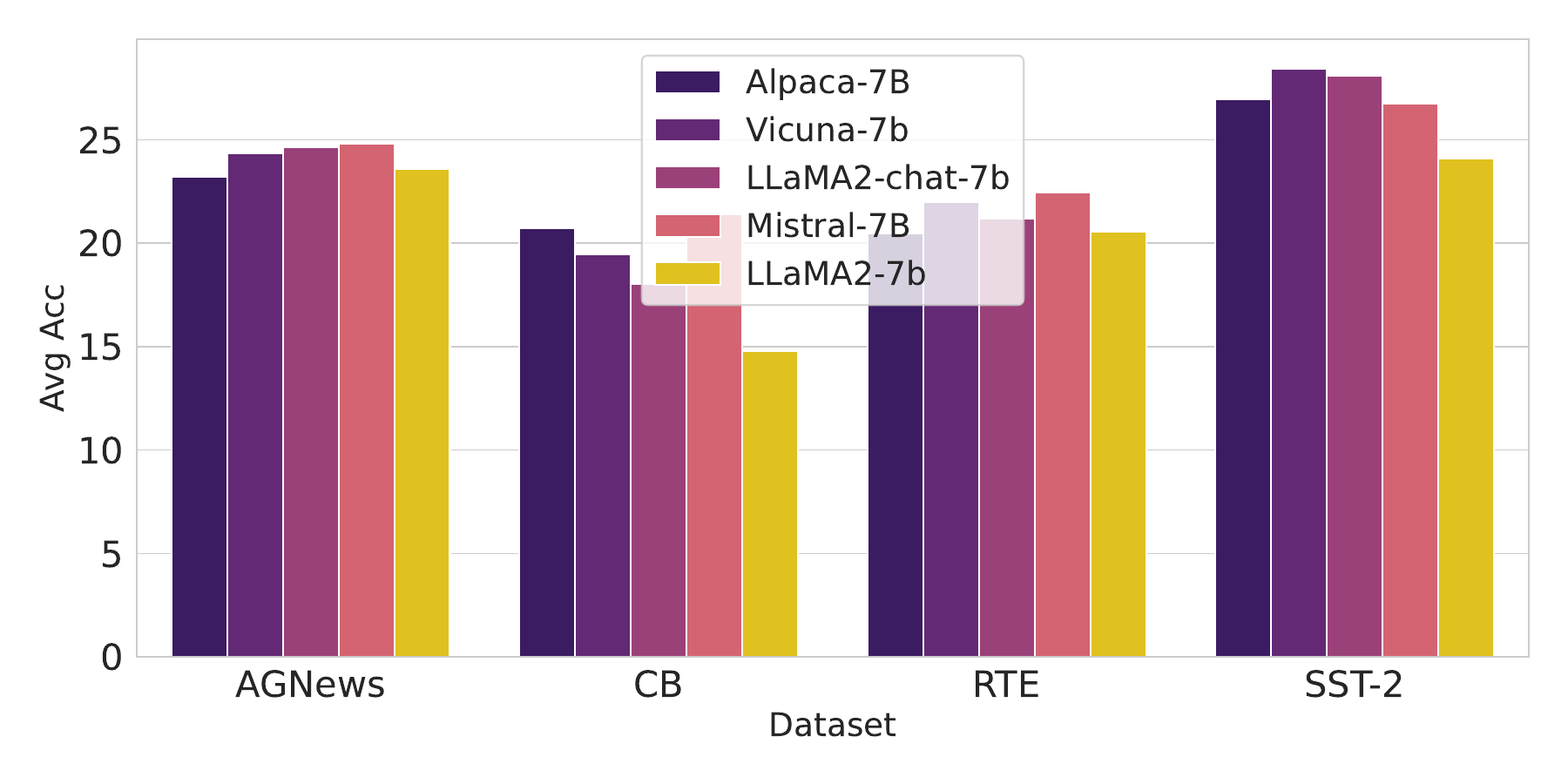} 
\label{fig:icl-finetune}
}
\subfigure[Calibration error]{
\includegraphics[width=.48\textwidth]{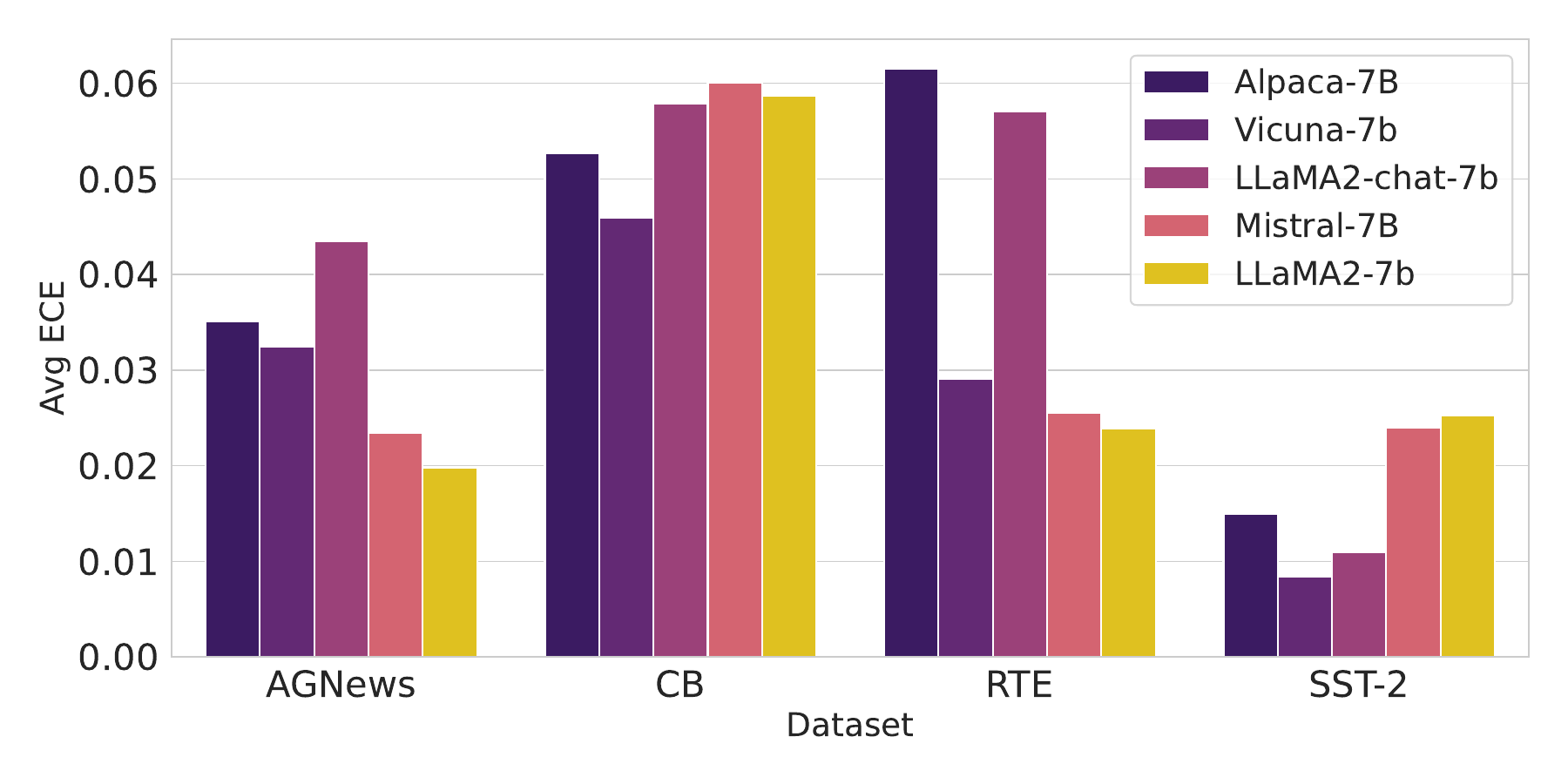} 
}
\end{minipage}
\caption{Accuracy and calibration errors of base models LLaMA and Mistral, as well as fine-tuned variants. Reported Acc and ECE results are averaged across experiments conducted with $\{0, 1, 2, 4, 8\}$ shots.}
\label{fig:finetuned}
\end{figure*}

\begin{table}[]
\centering
\resizebox{\linewidth}{!}{%
\begin{tabular}{ccccccccc}
\toprule \hline
Dataset & Strategy
& 0-shot & 1-shot & 2-shot & 3-shot & 4-shot & 8-shot
& Avg %
   \\ \midrule
\multirow{4}{*}{SST-2} & None & 0.043 & 0.223 & 0.119 & 0.101 & 0.060 & 0.049 & 0.099 \\
 & $0$-shot & 0.043 & 0.216 & \textbf{0.082} & 0.074 & 0.047 & 0.057 & 0.087 \\
 & $k$-shot & \textbf{0.034} & 0.197 & 0.101 & 0.079 & \textbf{0.041} & \textbf{0.038} & 0.139 \\
 & Fix $w$ & 0.035 & \textbf{0.176} & 0.086 & \textbf{0.073} & 0.047 & 0.043 & \textbf{0.077} \\
 \hline
\multirow{4}{*}{CB}  & None & 0.125 & 0.316 & 0.177 & 0.202 & 0.221 & 0.210 & 0.203 \\
 & $0$-shot & \textbf{0.015} & 0.252 & 0.162 & 0.217 & 0.217 & 0.199 & 0.209 \\
 & $k$-shot & \textbf{0.015} & 0.357 & 0.187 & 0.188 & 0.212 & 0.216 & 0.214 \\
 & Fix $w$ & \textbf{0.015} & \textbf{0.217} & \textbf{0.159} & \textbf{0.173} & \textbf{0.182} & 0.210 & \textbf{0.190} \\
 \hline
\multirow{4}{*}{RTE} & None & 0.108 & 0.110 & 0.142 & 0.122 & 0.128 & 0.120 & 0.122 \\
 & $0$-shot & 0.107 & 0.112 & 0.143 & 0.114 & 0.125 & 0.116 & 0.119 \\
 & $k$-shot & 0.108 & 0.115 & 0.136 & 0.112 & 0.126 & 0.125 & 0.120 \\
 & Fix $w$ & \textbf{0.101} & \textbf{0.082} & \textbf{0.097} & \textbf{0.068} & \textbf{0.076} & \textbf{0.097} & \textbf{0.088} \\ \hline
\multirow{4}{*}{AGNews} & None & 0.089 & 0.057 & 0.071 & 0.121 & 0.085 & 0.123 & 0.090 \\
 & $0$-shot & \textbf{0.067} & 0.087 & 0.098 & 0.160 & 0.107 & 0.130 & 0.114 \\
 & $k$-shot & 0.083 & 0.074 & \textbf{0.059} & 0.109 & \textbf{0.073} & 0.082 & \textbf{0.079} \\
 & Fix $w$ & 0.080 & 0.074 & 0.080 & \textbf{0.091} & \textbf{0.073} & \textbf{0.080} & 0.080 \\ \hline \bottomrule
\end{tabular}%
}
\caption{\textbf{ECE for different calibration strategies using temperature scaling~\citep{guo2017calibration}} of base models LLaMA-2-7B across various shot settings.} \label{tab:temp}
\end{table}

\begin{table}[]
\centering
\resizebox{\linewidth}{!}{%
\begin{tabular}{ccccccccc}
\toprule \hline
Dataset & Strategy & 0-shot & 1-shot & 2-shot & 3-shot & 4-shot & 8-shot & Avg  \\ \midrule
\multirow{4}{*}{SST-2} & None & 0.043 & 0.223 & 0.119 & 0.101 & 0.060 & 0.049 & 0.099 \\
 & $0$-shot & \textbf{0.015} & 0.062 & 0.055 & 0.060 & 0.062 & 0.057 & 0.052 \\
 & $k$-shot & 0.022 & 0.007 & \textbf{0.008} & \textbf{0.013} & \textbf{0.004} & \textbf{0.008} & \textbf{0.010} \\
 & Fix $w$ & 0.021 & \textbf{0.004} & \textbf{0.008} & \textbf{0.010} & 0.005 & 0.009 & \textbf{0.010} \\
 \hline
\multirow{4}{*}{CB} & None & 0.125 & 0.316 & 0.177 & 0.202 & 0.221 & 0.210 & 0.203 \\
 & $0$-shot & 0.122 & 0.130 & 0.121 & \textbf{0.086} & \textbf{0.083} & 0.119 & 0.110 \\
 & $k$-shot & \textbf{0.119} & 0.109 & 0.100 & 0.109 & 0.101 & \textbf{0.049} & \textbf{0.094} \\
 & Fix $w$& 0.119 & \textbf{0.088} & \textbf{0.085} & 0.110 & 0.121 & 0.069 & 0.099 \\
 \hline
\multirow{4}{*}{RTE} & None & 0.108 & 0.110 & 0.142 & 0.122 & 0.128 & 0.120 & 0.122 \\
 & $0$-shot & 0.078 & \textbf{0.083} & 0.090 & 0.100 & \textbf{0.102} & 0.115 & \textbf{0.093} \\
 & $k$-shot & 0.089 & 0.084 & \textbf{0.089} & \textbf{0.095} & 0.101 & \textbf{0.112} & 0.096 \\
 & Fix $w$& \textbf{0.077} & 0.086 & 0.092 & 0.100 & 0.108 & 0.117 & 0.099 \\ \hline
\multirow{4}{*}{AGNews} & None & 0.089 & 0.057 & 0.071 & 0.121 & 0.085 & 0.123 & 0.090 \\
 & $0$-shot & 0.007 & 0.013 & \textbf{0.011} & 0.014 & 0.013 & 0.014 & 0.013 \\
 & $k$-shot & \textbf{0.001} & \textbf{0.009} & 0.015 & 0.018 & \textbf{0.005} & \textbf{0.005} & \textbf{0.009} \\
 & Fix $w$& 0.015 & 0.019 & 0.019 & \textbf{0.005} & 0.008 & 0.017 & 0.013 \\ \hline \bottomrule
\end{tabular}%
}
\caption{\textbf{ECE for different calibration strategies using scaling-binning~\citep{kumar2019verified} calibrator} of base models LLaMA-2-7B across various shot settings.} \label{tab:scal_bin}
\end{table}

\begin{figure*}[h]
\begin{minipage}[h]{\textwidth}
\centering
\subfigure[$0$-shot]{
\includegraphics[width=.22\textwidth]{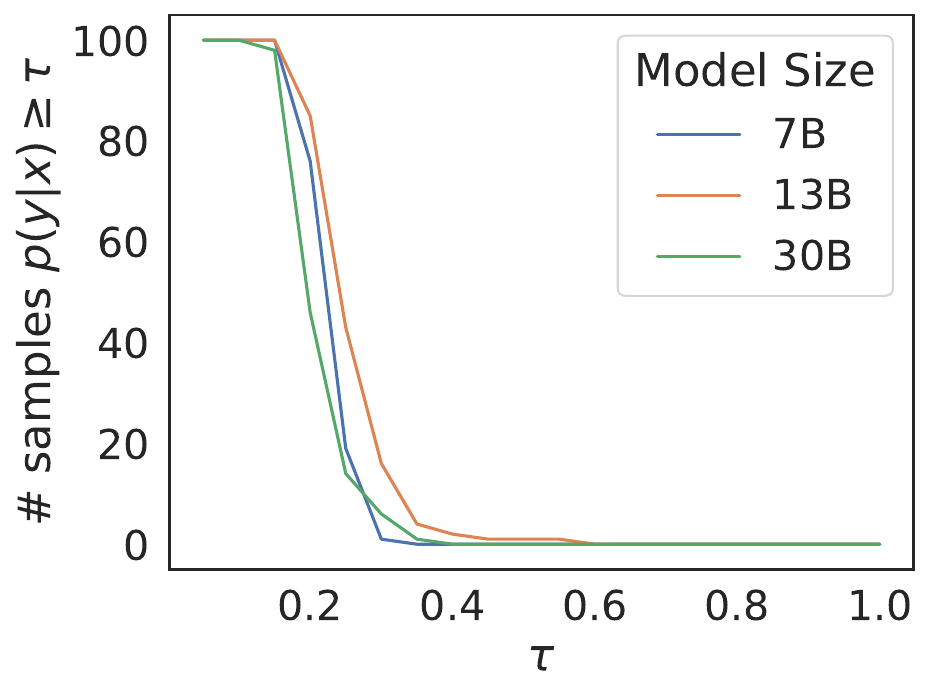}
\label{fig:gap-ece-0shot}
}
\subfigure[$1$-shot]{
\includegraphics[width=.22\textwidth]{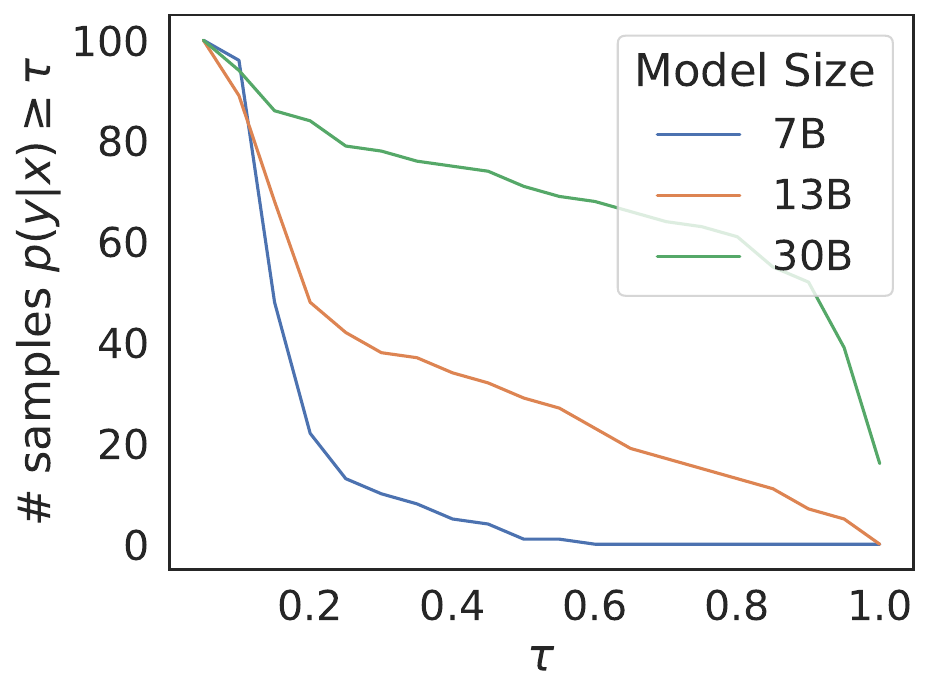} 
\label{fig:gap-loss} 
}
\subfigure[$4$-shot]{
\includegraphics[width=.22\textwidth]{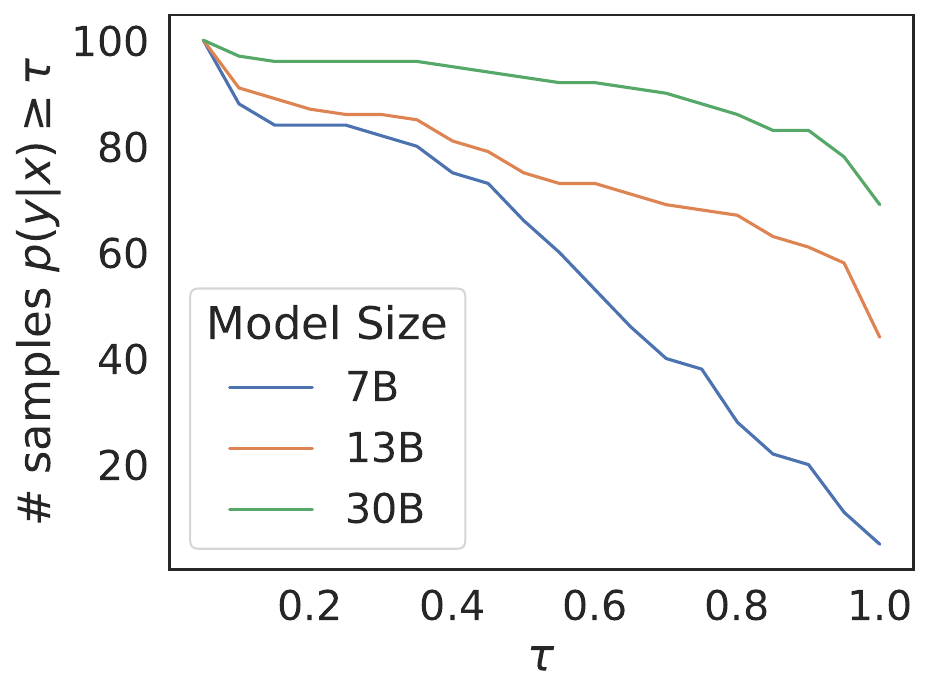} 
\label{fig:gap-loss-2} 
}
\subfigure[$8$-shot]{
\includegraphics[width=.22\textwidth]{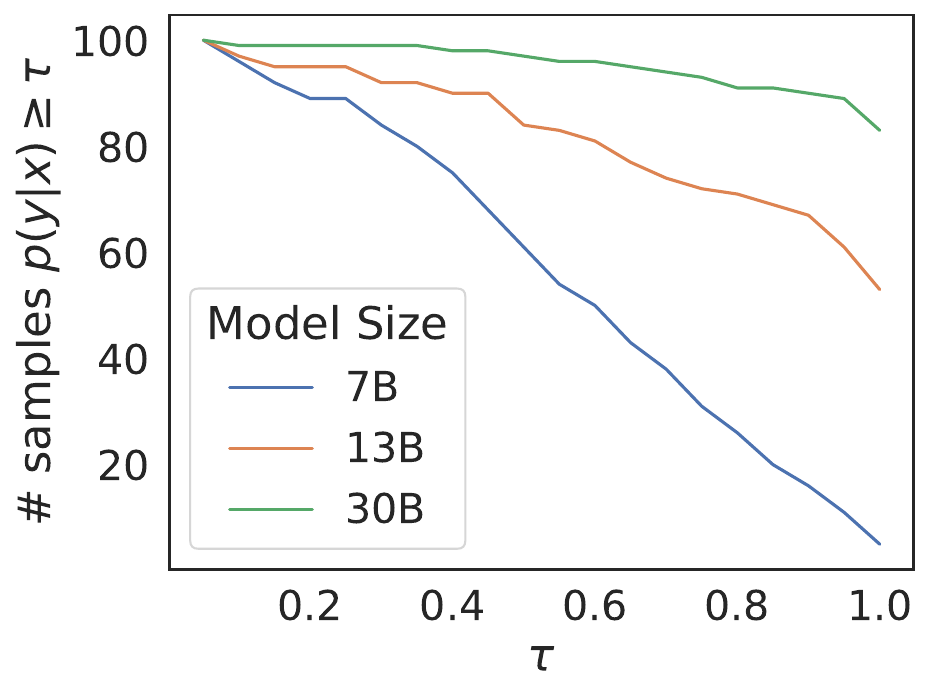} 
\label{fig:gap-loss-3} 
}
\end{minipage}
\caption{
\textbf{Illustration of confidence distribution.} The number of samples whose confidence is greater than a threshold on Commonsense QA. 
}
\label{fig:conf}
\end{figure*}

\begin{figure*}[h]
\begin{minipage}[h]{\textwidth}
\centering
\subfigure[$0$-shot]{
\includegraphics[width=.22\textwidth]{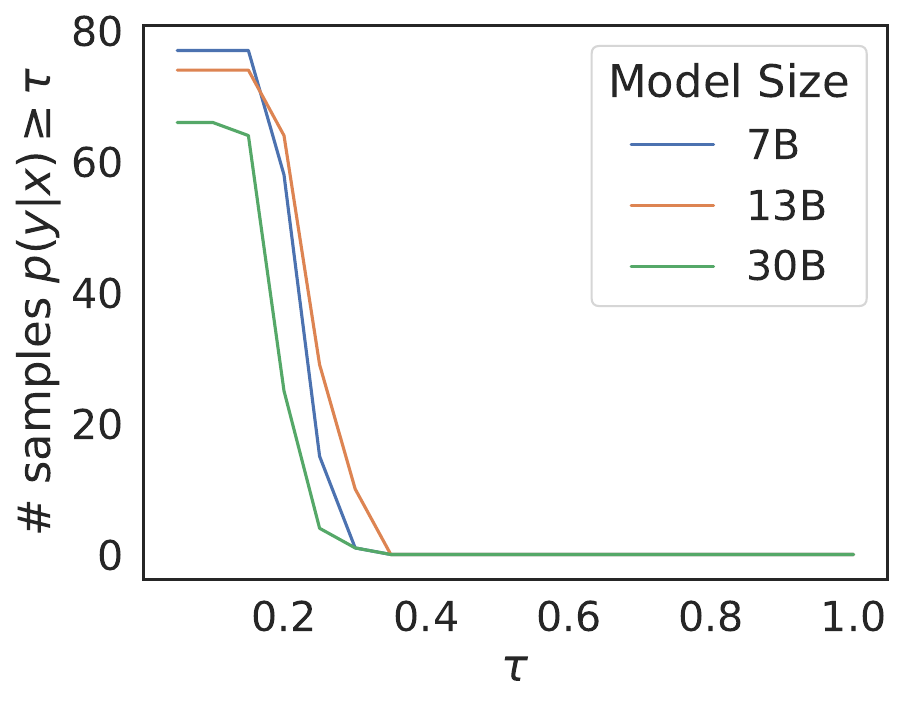}
\label{fig:gap-ece}
}
\subfigure[$1$-shot]{
\includegraphics[width=.22\textwidth]{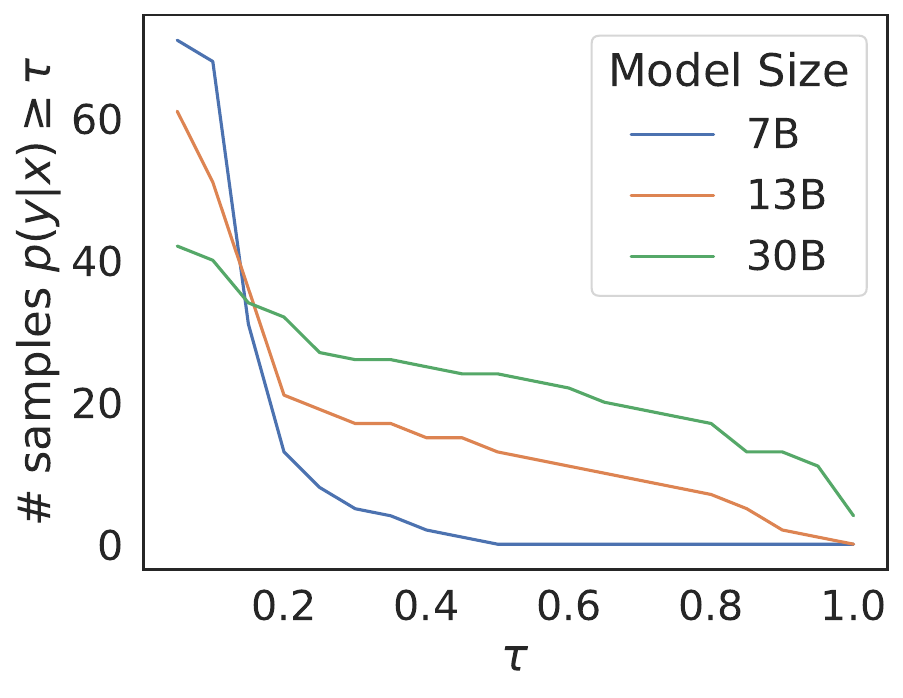} 
\label{fig:gap-loss-1shot} 
}
\subfigure[$4$-shot]{
\includegraphics[width=.22\textwidth]{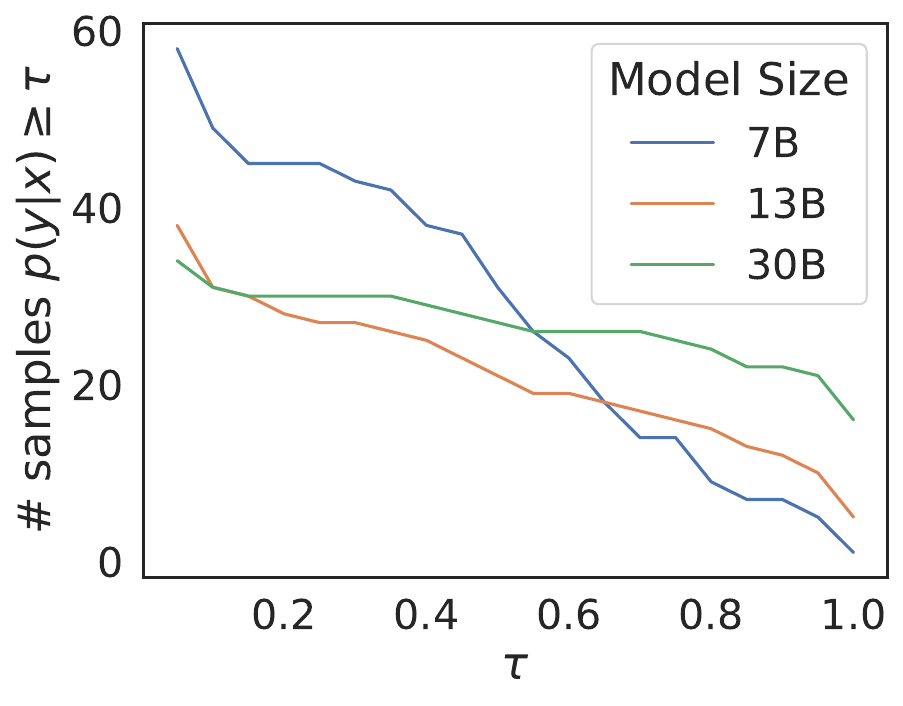} 
\label{fig:gap-loss-4shot} 
}
\subfigure[$8$-shot]{
\includegraphics[width=.22\textwidth]{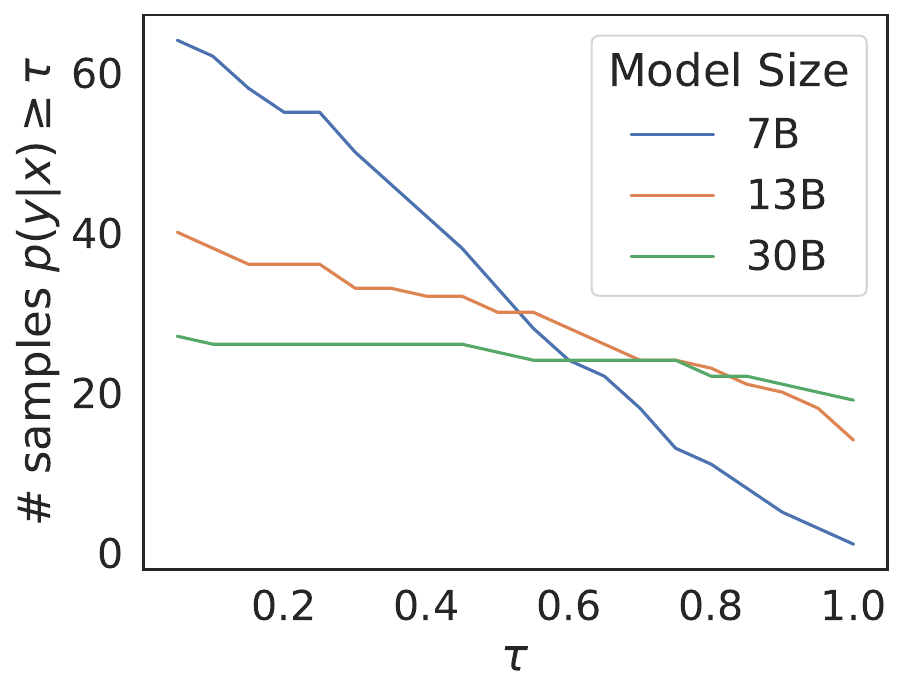} 
\label{fig:gap-loss-8shot} 
}
\end{minipage}
\caption{
The number of \textbf{wrongly classified} examples whose confidence is above a threshold with different numbers of shots on Commonsense QA.
}
\label{fig:wrong}
\end{figure*}

\noindent
\textbf{The effect of fine-tuning.}
We show that vicuna, alpaca, and LLaMA2-Chat are all more accurate but less calibrated than their LLaMA counterpart backbones (Fig. \ref{fig:finetuned}), the margin is especially large for reasoning tasks and vicuna. 
Our finding indicates that fine-tuning might significantly degrade calibration, corroborating the evidence reported in GPT-4 \citep{openai2023gpt4}, albeit it can greatly improve the reasoning accuracy.
Our results provide evidence that though fine-tuning on carefully curated datasets can greatly improve question-answering performance, especially for hard tasks like reasoning problems, attention may need to be paid when assessing the calibration of those models' predictions. 
Moreover, we include results of Mistral-7B \citep{jiang2023mistral}, a sparse Mixture of Experts (MoEs) architecture with sliding window attention. 
As a base model, it shows similar performance and calibration compared with LLaMA2-7B, indicating that our conclusion still holds for the model pre-trained with significantly different data and architecture. Comprehensive results and variance across different configurations are elaborated in Appendix \cref{tab:mean_std_baselines}.

\noindent
\textbf{The effect of prompt formats.} 
In our study, we explore the effects of different prompt strategies using three distinct methods. We consider predicting the label $y_n$ of test example $x_n$. 
First, the \textit{Repeat-context} approach involves constructing prompts as \( w_0 = [x_1, x_1, ..., x_1, y_1, x_n] \), where the context \( x_1 \) is repeated n-1 times, but the label \( y_1 \) is not included in the repetition. 
Next, the \textit{Repeat-prompt} strategy shapes the prompt as \( w_0 = [x_1, y_1, ..., x_1, y_1, x_n] \), where both the context \( x_1 \) and the label \( y_1 \) are repeated n-1 times. Finally, the \textit{Normal} involves constructing the prompt as \( w_0 = [x_1, y_1, x_2, y_2, ..., x_{n-1}, y_{n-1}, x_n] \), systematically incorporating distinct context-label pairs.

The findings, as detailed in Tab.~\ref{tab:prompt_rep}, reveal certain insights: 
firstly, integrating labels within prompts significantly decreases uncertainty and enhances learning performance. The reason may be that it aids the model in understanding the label space, which leads to better classification outcomes. In contrast, simply repeating the context without labels does not lead to better outcomes. 
Secondly, the diversity of ICL examples in the prompt greatly affects performance, a potential explanation is it promotes better task learning \citep{pan2023context}.
Those observations corroborate that ICL is performant when the number of ICL examples is large and they demonstrate consistent task properties. Importantly, the trade-off persists for different controlled scenarios, i.e. as we increase the number of ICL examples, models tend to be first more miscalibrated and then calibrated.

\begin{table*}[h]
\centering
\resizebox{0.98\textwidth}{!}{
\begin{tabular}{c|c|c|c|c|c|c|c|c|c|c|c|c}
\toprule \hline
\multirow{2}{*}{Dataset} & \multicolumn{12}{c}{LLaMA-30B} \\
\cline{2-13}
&  \multicolumn{4}{c|}{Norm} &  \multicolumn{4}{c|}{Entropy} &  \multicolumn{4}{c}{Confidence} \\
\cline{1-13}
& 0-shot & 1-shot & 4-shot & 8-shot & 0-shot & 1-shot & 4-shot & 8-shot & 0-shot & 1-shot & 4-shot & 8-shot \\
\hline
AGNews & 78.8   & 92.3  & 92.1  & 92.2 & 3.920	& 0.650 &	0.595 &	0.444  & 0.214 &	0.821 &	0.819 &	0.865 \\
\hline
CB & 88.4 	& 91.7 	& 89.2 	& 87.9 & 3.857	& 1.266 &	0.935 &	0.823 & 0.193 & 0.566 & 0.629 & 0.577  \\
\hline
DBPdia & 77.9&	89.5&	91.0&	90.1  & 4.105	 & 1.438 &	0.848	 & 0.718 & 0.078 &	0.578 &	0.705 &	0.671  \\
\bottomrule 
\end{tabular}
}
\caption{\textbf{Norm of representation, entropy, and confidence} of LLaMA-30B across three text classification datasets.}
\label{tab:norm}
\end{table*}

\subsection{Qualitative Results}

\textbf{Reliability diagram and confidence histogram.} A reliability diagram is a graphical tool used to evaluate the calibration of probabilistic predictions of a model across multiple classes; it compares the predicted probabilities of each class against the actual outcomes, with a perfectly calibrated model having its values lie on the diagonal $y=x$ line. A confidence histogram, on the other hand, displays the distribution of the model's prediction confidences across all classes, showing how often the model predicts certain probabilities.

Recall that we found significant miscalibration for reasoning with CoT settings, therefore we closely examine the poorly calibrated reasoning cases using the above plots (Fig. \ref{fig:reliability} and Fig. \ref{fig:reliability_appendix}). 
Our results on 4-shot settings show that for the reasoning problems (Strategy QA, Commonsense QA, OpenBook QA, World Tree) we consider, models are consistently over-confident with ECEs above 0.15. 
Larger models are better both in both ACC and ECE but for OpenBook QA, calibration worsens as the model size increases. 
Moreover, it's observed that confidence scores tend to concentrate on high values as we enlarge the model size. 
Especially in Commonsense QA and OpenBook QA, the confidence level of nearly all predictions of 13B and 30B models predominantly exceeds 0.8.

\subsection{Ablation Studies}

For case studies, we research how miscalibration can impact the selective classification of LMs, where models are supposed to abstain from uncertain predictions in high-stakes settings.

\noindent
\textbf{Ablation with model sizes.}
As we enlarge the size of models, they become more confident (as measured by the confidence histogram) and accurate (Fig. \ref{fig:reliability}). 
Moreover, the ECE first increases and then decreases. In some settings like SST-2 and OpenBookQA, calibration errors may have a negative correlation with model sizes (Appendix Tab.\ref{tab:main_cot}).

\begin{table*}[h]
    \centering
    \resizebox{0.8\textwidth}{!}{%
    \hspace*{-0.3cm}\begin{tabularx}{\textwidth}{|c|c|X|c|c|}
        \cline{1-5}
        Dataset & Case & \centering Reasoning & $\hat{y}$ & $p$ \\
        \cline{1-5}
        \multirow{12}{*}{Commonsense} & \multicolumn{4}{c|}{Question: What would you do if you have curiosity but are blind and paralyzed?} \\
        \cline{2-5}
        & \multicolumn{4}{c|}{Options: A) hear news, B) analyse, C) go somewhere, D) examine thing, E) see favorite show} \\
        \cline{2-5}
        & \multicolumn{4}{c|}{Ground Truth: B} \\
        \cline{2-5}
        & 7B & If you have curiosity but are blind and paralyzed, you would go somewhere to see things. You cannot hear news, analyse, go somewhere, examine thing or see favorite show. & C & 0.5046 \\
        \cline{2-5}
        & 13B & If you are blind and paralyzed, you can't see anything. So, you can't go somewhere. You can't examine thing. You can't see favorite show. You can't analyse. So, you can hear news. & A & 0.8684 \\
        \cline{2-5}
        & 30B & If you are blind and paralyzed, you cannot see your favorite show. You cannot hear news as you are blind. You cannot examine thing as you are paralyzed. You can go somewhere to see things. & C & 0.9720 \\
        \cline{1-5}
        \multirow{6}{*}{Open Book} & \multicolumn{4}{c|}{Question: What would you do if you have curiosity but are blind and paralyzed?} \\
        \cline{2-5}
        & \multicolumn{4}{c|}{Options: A) hear news, B) analyse, C) go somewhere, D) examine thing, E) see favorite show} \\
        \cline{2-5}
        & \multicolumn{4}{c|}{Ground Truth: B} \\
        \cline{2-5}
        & 7B & A girl and her mom have the same hair length. & D & 0.6365 \\
        \cline{2-5}
        & 13B & A girl and her mom have the same date of birth. & A & 0.9782 \\
        \cline{2-5}
        & 30B & A girl and her mom have the same genes. & A & 0.9831 \\
        \cline{2-5} \hline
    \end{tabularx}
    }
    \caption{Qualitative Results of LLaMA on Commonsense and OpenBook}
    \label{tab:confident_wrong}
\end{table*}

\noindent
\textbf{Examine the hidden state and confidence score.}
To better understand the miscalibration issue of ICL, 
we conduct fined-grained experiments to examine ICL properties: 
we measure the norm of the representation vectors\footnote{The representation vector refers to the intermediate representation before the linear prediction layer.} for different numbers of shots in ICL. %
Meanwhile, we also measure the confidence and entropy of the prediction for $y_n$, and the results are summarized in  Tab.~\ref{tab:norm}.
When switching from 0-shot to 1-shot, all three measurements (representation norm, entropy, and confidence) drastically change;
on the other hand, when $k$ increases ($1\rightarrow4\rightarrow8$), the change of measures would become smoother. 
Our discovery shows that adding in-context examples can substantially impact model behaviors while the model behaves relatively similarly for various shots once the task is specified ($k\neq0)$. 
Meanwhile, more ICL samples lead to smaller entropy and higher confidence in most cases. 
Considering the alterations in feature representation, which can manifest in either an augmentation of the representation's norm or a shift in direction, quantifying changes in feature direction poses challenges. Thus, we have chosen to examine changes in the norm as a surrogate measure,
suggesting that as the number of ICL samples increases, there is a systematic alteration in the model's features.

\noindent
\textbf{Confidence and wrongly classified reasoning examples.}
To inspect the failure modes of LMs, we randomly sample 100 reasoning examples of LLaMA and plot the distribution of wrongly predicted samples and the confidence scores via thresholding. 
Similar to previous observations, as model sizes and the number of ICL examples scale up, LMs would generate more confident samples (Fig. \ref{fig:conf} (c, d)).
We observe behaviors where models with larger sizes may be more error-prone and tend to generate more confidently wrong explanatory samples (Fig. \ref{fig:wrong}). 

\noindent
\textbf{Examples of hallucinated explanations for highly confident predictions.}
Next, we showcase in Tab. \ref{tab:confident_wrong} that models generate both wrong explanations and incorrect predictions with high confidence. We also observe that most of the wrong predictions are highly confident, thus we manually examine the correctness of explanations on commonsense QA, and found its high correlations with predicted answer accuracy, which is the opposite of token-level explainability that tends to get worse when the accuracy improves. For additional qualitative examination of LLaMA's performance on Strategy QA and WorldTree, please refer to Table \ref{tab:confident_wrong_app}.

\section{Discussion and Concluding Remarks}
\label{sec:conclusion}
In our investigation of the token-level calibration of in-context learning in contemporary language models, we illustrate the intricate trade-off between ICL performance and calibration. 
Our findings underscore the importance of being circumspect in model deployment, as maximizing ICL performance does not invariably translate to improved calibration for low-shot and reasoning settings. 
As LMs continue to evolve and gain more capabilities such as having long enough context windows that can include the whole training set as in-context examples for some downstream tasks, our result can be pedagogical when users would like to examine their uncertainty through prediction probabilities. Moreover, the work suggests the following future directions:

\paragraph{Calibration beyond classification regimes.}
Our findings indicate that in multi-choice or multi-class classification tasks, even though the calibration of answer tokens may deteriorate in high-performance settings, there may be a positive correlation between accuracy and the correctness of explanations in reasoning tasks. 
This suggests potential avenues for future research in exploring strategies such as the use of hedging words to express uncertainty and examining their relationship with predictive performance.

\paragraph{Implications in assessing beliefs of LMs.}

Previous works show that the expected calibration error would decrease monotonically as the number of ICL examples increases \citep{kadavath2022language} when querying LMs for answer probabilities. 
However, we find that zero-shot performance might be weak for models less than 30B, and in low-shot settings, calibration errors can sometimes be even worse than zero-shot. 
This implies that a close examination and careful control of epistemic uncertainty and aleatoric uncertainty can be needed before deriving conclusions in truthfulness \citep{liu2023cognitive, azaria2023internal} for low-shot settings.

\noindent
\textbf{Limitations}. 
We acknowledge the need to expand our evaluation, which is primarily focused on QA and classification tasks, beyond existing open-sourced language models and datasets. 
Moreover, we didn't consider nuances such as inherent disagreement about labels \citep{baan2022stop} and adaptive calibration error measures \citep{nixon2019measuring} that might be important in certain use cases: it's worth noting that situations may arise where multiple labels share the highest predicted probability. In such instances, the definition (Eq. (\ref{eq:def})) doesn't automatically become false; instead, we opt for the first maximal probability. These cases are less likely to occur in most of our experimental setups, where a substantial margin consistently exists between different labels.

\subsection*{Acknowledgment} 
We thank Yu Bai, David Childers, Jean-Stanislas Denain for their valuable feedback.
HZ is supported by an Eric and Susan Dunn Graduate Fellowship.
YY acknowledges support from the joint Simons Foundation-NSF DMS grant \#2031899. 
SK acknowledges support from the Office of Naval Research under award N00014-22-1-2377 and the National Science Foundation Grant under award \#IIS 2229881. 
Kempner Institute computing resources enabled this work.
This work has been made possible in part by a gift from the Chan Zuckerberg Initiative Foundation to establish the Kempner Institute for the Study of Natural and Artificial Intelligence.

\bibliography{ref}

\begin{thebibliography}{69}
\expandafter\ifx\csname natexlab\endcsname\relax\def\natexlab#1{#1}\fi

\bibitem[{Aky{\"u}rek et~al.(2022)Aky{\"u}rek, Schuurmans, Andreas, Ma, and
  Zhou}]{akyurek2022learning}
Ekin Aky{\"u}rek, Dale Schuurmans, Jacob Andreas, Tengyu Ma, and Denny Zhou.
  2022.
\newblock What learning algorithm is in-context learning? investigations with
  linear models.
\newblock \emph{arXiv preprint arXiv:2211.15661}.

\bibitem[{Angelopoulos et~al.(2022)Angelopoulos, Bates, Fisch, Lei, and
  Schuster}]{angelopoulos2022conformal}
Anastasios~N Angelopoulos, Stephen Bates, Adam Fisch, Lihua Lei, and Tal
  Schuster. 2022.
\newblock Conformal risk control.
\newblock \emph{arXiv preprint arXiv:2208.02814}.

\bibitem[{Azaria and Mitchell(2023)}]{azaria2023internal}
Amos Azaria and Tom Mitchell. 2023.
\newblock The internal state of an llm knows when its lying.
\newblock \emph{arXiv preprint arXiv:2304.13734}.

\bibitem[{Baan et~al.(2022)Baan, Aziz, Plank, and Fernandez}]{baan2022stop}
Joris Baan, Wilker Aziz, Barbara Plank, and Raquel Fernandez. 2022.
\newblock Stop measuring calibration when humans disagree.
\newblock \emph{arXiv preprint arXiv:2210.16133}.

\bibitem[{Bai et~al.(2023)Bai, Chen, Wang, Xiong, and
  Mei}]{bai2023transformers}
Yu~Bai, Fan Chen, Huan Wang, Caiming Xiong, and Song Mei. 2023.
\newblock Transformers as statisticians: Provable in-context learning with
  in-context algorithm selection.
\newblock \emph{arXiv preprint arXiv:2306.04637}.

\bibitem[{Bansal et~al.(2021)Bansal, Wu, Zhou, Fok, Nushi, Kamar, Ribeiro, and
  Weld}]{bansal2021does}
Gagan Bansal, Tongshuang Wu, Joyce Zhou, Raymond Fok, Besmira Nushi, Ece Kamar,
  Marco~Tulio Ribeiro, and Daniel Weld. 2021.
\newblock Does the whole exceed its parts? the effect of ai explanations on
  complementary team performance.
\newblock In \emph{Proceedings of the 2021 CHI Conference on Human Factors in
  Computing Systems}, pages 1--16.

\bibitem[{Bhatt et~al.(2021)Bhatt, Antor{\'a}n, Zhang, Liao, Sattigeri,
  Fogliato, Melan{\c{c}}on, Krishnan, Stanley, Tickoo
  et~al.}]{bhatt2021uncertainty}
Umang Bhatt, Javier Antor{\'a}n, Yunfeng Zhang, Q~Vera Liao, Prasanna
  Sattigeri, Riccardo Fogliato, Gabrielle Melan{\c{c}}on, Ranganath Krishnan,
  Jason Stanley, Omesh Tickoo, et~al. 2021.
\newblock Uncertainty as a form of transparency: Measuring, communicating, and
  using uncertainty.
\newblock In \emph{Proceedings of the 2021 AAAI/ACM Conference on AI, Ethics,
  and Society}, pages 401--413.

\bibitem[{Braverman et~al.(2020)Braverman, Chen, Kakade, Narasimhan, Zhang, and
  Zhang}]{braverman2020calibration}
Mark Braverman, Xinyi Chen, Sham Kakade, Karthik Narasimhan, Cyril Zhang, and
  Yi~Zhang. 2020.
\newblock Calibration, entropy rates, and memory in language models.
\newblock In \emph{International Conference on Machine Learning}, pages
  1089--1099. PMLR.

\bibitem[{Brown et~al.(2020)Brown, Mann, Ryder, Subbiah, Kaplan, Dhariwal,
  Neelakantan, Shyam, Sastry, Askell et~al.}]{brown2020language}
Tom Brown, Benjamin Mann, Nick Ryder, Melanie Subbiah, Jared~D Kaplan, Prafulla
  Dhariwal, Arvind Neelakantan, Pranav Shyam, Girish Sastry, Amanda Askell,
  et~al. 2020.
\newblock Language models are few-shot learners.
\newblock \emph{Advances in neural information processing systems},
  33:1877--1901.

\bibitem[{Bu{\c{c}}inca et~al.(2021)Bu{\c{c}}inca, Malaya, and
  Gajos}]{buccinca2021trust}
Zana Bu{\c{c}}inca, Maja~Barbara Malaya, and Krzysztof~Z Gajos. 2021.
\newblock To trust or to think: cognitive forcing functions can reduce
  overreliance on ai in ai-assisted decision-making.
\newblock \emph{Proceedings of the ACM on Human-Computer Interaction},
  5(CSCW1):1--21.

\bibitem[{Camburu et~al.(2018)Camburu, Rockt{\"a}schel, Lukasiewicz, and
  Blunsom}]{camburu2018snli}
Oana-Maria Camburu, Tim Rockt{\"a}schel, Thomas Lukasiewicz, and Phil Blunsom.
  2018.
\newblock e-snli: Natural language inference with natural language
  explanations.
\newblock \emph{Advances in Neural Information Processing Systems}, 31.

\bibitem[{Chowdhery et~al.(2023)Chowdhery, Narang, Devlin, Bosma, Mishra,
  Roberts, Barham, Chung, Sutton, Gehrmann et~al.}]{chowdhery2023palm}
Aakanksha Chowdhery, Sharan Narang, Jacob Devlin, Maarten Bosma, Gaurav Mishra,
  Adam Roberts, Paul Barham, Hyung~Won Chung, Charles Sutton, Sebastian
  Gehrmann, et~al. 2023.
\newblock Palm: Scaling language modeling with pathways.
\newblock \emph{Journal of Machine Learning Research}, 24(240):1--113.

\bibitem[{Cole et~al.(2023)Cole, Zhang, Gillick, Eisenschlos, Dhingra, and
  Eisenstein}]{cole2023selectively}
Jeremy~R Cole, Michael~JQ Zhang, Daniel Gillick, Julian~Martin Eisenschlos,
  Bhuwan Dhingra, and Jacob Eisenstein. 2023.
\newblock Selectively answering ambiguous questions.
\newblock \emph{arXiv preprint arXiv:2305.14613}.

\bibitem[{Dawid(1982)}]{dawid1982well}
A~Philip Dawid. 1982.
\newblock The well-calibrated bayesian.
\newblock \emph{Journal of the American Statistical Association},
  77(379):605--610.

\bibitem[{DeGroot and Fienberg(1983)}]{degroot1983comparison}
Morris~H DeGroot and Stephen~E Fienberg. 1983.
\newblock The comparison and evaluation of forecasters.
\newblock \emph{Journal of the Royal Statistical Society: Series D (The
  Statistician)}, 32(1-2):12--22.

\bibitem[{Desai and Durrett(2020)}]{desai2020calibration}
Shrey Desai and Greg Durrett. 2020.
\newblock Calibration of pre-trained transformers.
\newblock \emph{arXiv preprint arXiv:2003.07892}.

\bibitem[{Dubois et~al.(2023)Dubois, Li, Taori, Zhang, Gulrajani, Ba, Guestrin,
  Liang, and Hashimoto}]{dubois2023alpacafarm}
Yann Dubois, Xuechen Li, Rohan Taori, Tianyi Zhang, Ishaan Gulrajani, Jimmy Ba,
  Carlos Guestrin, Percy Liang, and Tatsunori~B. Hashimoto. 2023.
\newblock \href {http://arxiv.org/abs/2305.14387} {Alpacafarm: A simulation
  framework for methods that learn from human feedback}.

\bibitem[{Fei et~al.(2023)Fei, Hou, Chen, and Bosselut}]{fei2023mitigating}
Yu~Fei, Yifan Hou, Zeming Chen, and Antoine Bosselut. 2023.
\newblock Mitigating label biases for in-context learning.
\newblock \emph{arXiv preprint arXiv:2305.19148}.

\bibitem[{Fisch et~al.(2020)Fisch, Schuster, Jaakkola, and
  Barzilay}]{fisch2020efficient}
Adam Fisch, Tal Schuster, Tommi Jaakkola, and Regina Barzilay. 2020.
\newblock Efficient conformal prediction via cascaded inference with expanded
  admission.
\newblock \emph{arXiv preprint arXiv:2007.03114}.

\bibitem[{Garg et~al.(2022)Garg, Tsipras, Liang, and Valiant}]{garg2022can}
Shivam Garg, Dimitris Tsipras, Percy~S Liang, and Gregory Valiant. 2022.
\newblock What can transformers learn in-context? a case study of simple
  function classes.
\newblock \emph{Advances in Neural Information Processing Systems},
  35:30583--30598.

\bibitem[{Geva et~al.(2021)Geva, Khashabi, Segal, Khot, Roth, and
  Berant}]{geva2021did}
Mor Geva, Daniel Khashabi, Elad Segal, Tushar Khot, Dan Roth, and Jonathan
  Berant. 2021.
\newblock Did aristotle use a laptop? a question answering benchmark with
  implicit reasoning strategies.
\newblock \emph{Transactions of the Association for Computational Linguistics},
  9:346--361.

\bibitem[{Gonz{\'a}lez et~al.(2021)Gonz{\'a}lez, Bansal, Fan, Mehdad, Jia, and
  Iyer}]{gonzalez2021explanations}
Ana~Valeria Gonz{\'a}lez, Gagan Bansal, Angela Fan, Yashar Mehdad, Robin Jia,
  and Srinivasan Iyer. 2021.
\newblock Do explanations help users detect errors in open-domain qa? an
  evaluation of spoken vs. visual explanations.
\newblock In \emph{Findings of the Association for Computational Linguistics:
  ACL-IJCNLP 2021}, pages 1103--1116.

\bibitem[{Guo et~al.(2017)Guo, Pleiss, Sun, and
  Weinberger}]{guo2017calibration}
Chuan Guo, Geoff Pleiss, Yu~Sun, and Kilian~Q Weinberger. 2017.
\newblock On calibration of modern neural networks.
\newblock In \emph{International conference on machine learning}, pages
  1321--1330. PMLR.

\bibitem[{Han et~al.(2023)Han, Hao, Dong, Sun, and Wei}]{han2023prototypical}
Zhixiong Han, Yaru Hao, Li~Dong, Yutao Sun, and Furu Wei. 2023.
\newblock Prototypical calibration for few-shot learning of language models.
\newblock In \emph{The Eleventh International Conference on Learning
  Representations}.

\bibitem[{Jansen et~al.(2018)Jansen, Wainwright, Marmorstein, and
  Morrison}]{jansen2018worldtree}
Peter~A Jansen, Elizabeth Wainwright, Steven Marmorstein, and Clayton~T
  Morrison. 2018.
\newblock Worldtree: A corpus of explanation graphs for elementary science
  questions supporting multi-hop inference.
\newblock \emph{arXiv preprint arXiv:1802.03052}.

\bibitem[{Jiang et~al.(2023)Jiang, Sablayrolles, Mensch, Bamford, Chaplot,
  Casas, Bressand, Lengyel, Lample, Saulnier et~al.}]{jiang2023mistral}
Albert~Q Jiang, Alexandre Sablayrolles, Arthur Mensch, Chris Bamford,
  Devendra~Singh Chaplot, Diego de~las Casas, Florian Bressand, Gianna Lengyel,
  Guillaume Lample, Lucile Saulnier, et~al. 2023.
\newblock Mistral 7b.
\newblock \emph{arXiv preprint arXiv:2310.06825}.

\bibitem[{Jiang et~al.(2021)Jiang, Araki, Ding, and Neubig}]{jiang2021can}
Zhengbao Jiang, Jun Araki, Haibo Ding, and Graham Neubig. 2021.
\newblock How can we know when language models know? on the calibration of
  language models for question answering.
\newblock \emph{Transactions of the Association for Computational Linguistics},
  9:962--977.

\bibitem[{Kadavath et~al.(2022)Kadavath, Conerly, Askell, Henighan, Drain,
  Perez, Schiefer, Hatfield-Dodds, DasSarma, Tran-Johnson
  et~al.}]{kadavath2022language}
Saurav Kadavath, Tom Conerly, Amanda Askell, Tom Henighan, Dawn Drain, Ethan
  Perez, Nicholas Schiefer, Zac Hatfield-Dodds, Nova DasSarma, Eli
  Tran-Johnson, et~al. 2022.
\newblock Language models (mostly) know what they know.
\newblock \emph{arXiv preprint arXiv:2207.05221}.

\bibitem[{Kalai and Vempala(2023)}]{kalai2023calibrated}
Adam~Tauman Kalai and Santosh~S. Vempala. 2023.
\newblock \href {http://arxiv.org/abs/2311.14648} {Calibrated language models
  must hallucinate}.

\bibitem[{Kamath et~al.(2020)Kamath, Jia, and Liang}]{kamath2020selective}
Amita Kamath, Robin Jia, and Percy Liang. 2020.
\newblock Selective question answering under domain shift.
\newblock \emph{arXiv preprint arXiv:2006.09462}.

\bibitem[{Kuhn et~al.(2023)Kuhn, Gal, and Farquhar}]{kuhn2023semantic}
Lorenz Kuhn, Yarin Gal, and Sebastian Farquhar. 2023.
\newblock Semantic uncertainty: Linguistic invariances for uncertainty
  estimation in natural language generation.
\newblock \emph{arXiv preprint arXiv:2302.09664}.

\bibitem[{Kull and Flach(2015)}]{kull2015novel}
Meelis Kull and Peter Flach. 2015.
\newblock Novel decompositions of proper scoring rules for classification:
  Score adjustment as precursor to calibration.
\newblock In \emph{Machine Learning and Knowledge Discovery in Databases:
  European Conference, ECML PKDD 2015, Porto, Portugal, September 7-11, 2015,
  Proceedings, Part I 15}, pages 68--85. Springer.

\bibitem[{Kumar et~al.(2019)Kumar, Liang, and Ma}]{kumar2019verified}
Ananya Kumar, Percy~S Liang, and Tengyu Ma. 2019.
\newblock Verified uncertainty calibration.
\newblock \emph{Advances in Neural Information Processing Systems}, 32.

\bibitem[{Laban et~al.(2022)Laban, Schnabel, Bennett, and
  Hearst}]{laban2022summac}
Philippe Laban, Tobias Schnabel, Paul~N Bennett, and Marti~A Hearst. 2022.
\newblock Summac: Re-visiting nli-based models for inconsistency detection in
  summarization.
\newblock \emph{Transactions of the Association for Computational Linguistics},
  10:163--177.

\bibitem[{Lin et~al.(2022)Lin, Hilton, and Evans}]{lin2022teaching}
Stephanie Lin, Jacob Hilton, and Owain Evans. 2022.
\newblock Teaching models to express their uncertainty in words.
\newblock \emph{arXiv preprint arXiv:2205.14334}.

\bibitem[{Liu et~al.(2023)Liu, Casper, Hadfield-Menell, and
  Andreas}]{liu2023cognitive}
Kevin Liu, Stephen Casper, Dylan Hadfield-Menell, and Jacob Andreas. 2023.
\newblock \href {http://arxiv.org/abs/2312.03729} {Cognitive dissonance: Why do
  language model outputs disagree with internal representations of
  truthfulness?}

\bibitem[{Mielke et~al.(2022)Mielke, Szlam, Dinan, and
  Boureau}]{mielke2022reducing}
Sabrina~J Mielke, Arthur Szlam, Emily Dinan, and Y-Lan Boureau. 2022.
\newblock Reducing conversational agents’ overconfidence through linguistic
  calibration.
\newblock \emph{Transactions of the Association for Computational Linguistics},
  10:857--872.

\bibitem[{Mihaylov et~al.(2018)Mihaylov, Clark, Khot, and
  Sabharwal}]{OpenBookQA2018}
Todor Mihaylov, Peter Clark, Tushar Khot, and Ashish Sabharwal. 2018.
\newblock Can a suit of armor conduct electricity? a new dataset for open book
  question answering.
\newblock In \emph{EMNLP}.

\bibitem[{Nixon et~al.(2019)Nixon, Dusenberry, Zhang, Jerfel, and
  Tran}]{nixon2019measuring}
Jeremy Nixon, Michael~W Dusenberry, Linchuan Zhang, Ghassen Jerfel, and Dustin
  Tran. 2019.
\newblock Measuring calibration in deep learning.
\newblock In \emph{CVPR workshops}, volume~2.

\bibitem[{Nye et~al.(2021)Nye, Andreassen, Gur-Ari, Michalewski, Austin,
  Bieber, Dohan, Lewkowycz, Bosma, Luan et~al.}]{nye2021show}
Maxwell Nye, Anders~Johan Andreassen, Guy Gur-Ari, Henryk Michalewski, Jacob
  Austin, David Bieber, David Dohan, Aitor Lewkowycz, Maarten Bosma, David
  Luan, et~al. 2021.
\newblock Show your work: Scratchpads for intermediate computation with
  language models.
\newblock \emph{arXiv preprint arXiv:2112.00114}.

\bibitem[{Olsson et~al.(2022)Olsson, Elhage, Nanda, Joseph, DasSarma, Henighan,
  Mann, Askell, Bai, Chen et~al.}]{olsson2022context}
Catherine Olsson, Nelson Elhage, Neel Nanda, Nicholas Joseph, Nova DasSarma,
  Tom Henighan, Ben Mann, Amanda Askell, Yuntao Bai, Anna Chen, et~al. 2022.
\newblock In-context learning and induction heads.
\newblock \emph{arXiv preprint arXiv:2209.11895}.

\bibitem[{OpenAI(2023)}]{openai2023gpt4}
OpenAI. 2023.
\newblock Gpt-4 technical report.
\newblock \emph{https://cdn.openai.com/papers/gpt-4.pdf}.

\bibitem[{Pan et~al.(2023)Pan, Chan, Zou, Li, Basart, Woodside, Zhang, Emmons,
  and Hendrycks}]{pan2023rewards}
Alexander Pan, Jun~Shern Chan, Andy Zou, Nathaniel Li, Steven Basart, Thomas
  Woodside, Hanlin Zhang, Scott Emmons, and Dan Hendrycks. 2023.
\newblock Do the rewards justify the means? measuring trade-offs between
  rewards and ethical behavior in the machiavelli benchmark.
\newblock In \emph{International Conference on Machine Learning}, pages
  26837--26867. PMLR.

\bibitem[{Pan(2023)}]{pan2023context}
Jane Pan. 2023.
\newblock \emph{What In-Context Learning “Learns” In-Context: Disentangling
  Task Recognition and Task Learning}.
\newblock Ph.D. thesis, Princeton University.

\bibitem[{Petryk et~al.(2023)Petryk, Whitehead, Gonzalez, Darrell, Rohrbach,
  and Rohrbach}]{petryk2023simple}
Suzanne Petryk, Spencer Whitehead, Joseph~E Gonzalez, Trevor Darrell, Anna
  Rohrbach, and Marcus Rohrbach. 2023.
\newblock Simple token-level confidence improves caption correctness.
\newblock \emph{arXiv preprint arXiv:2305.07021}.

\bibitem[{Platt et~al.(1999)}]{platt1999probabilistic}
John Platt et~al. 1999.
\newblock Probabilistic outputs for support vector machines and comparisons to
  regularized likelihood methods.
\newblock \emph{Advances in large margin classifiers}, 10(3):61--74.

\bibitem[{Ravent{\'o}s et~al.(2023)Ravent{\'o}s, Paul, Chen, and
  Ganguli}]{raventos2023pretraining}
Allan Ravent{\'o}s, Mansheej Paul, Feng Chen, and Surya Ganguli. 2023.
\newblock Pretraining task diversity and the emergence of non-bayesian
  in-context learning for regression.
\newblock \emph{arXiv preprint arXiv:2306.15063}.

\bibitem[{Schick and Sch{\"u}tze(2021)}]{schick2021exploiting}
Timo Schick and Hinrich Sch{\"u}tze. 2021.
\newblock Exploiting cloze-questions for few-shot text classification and
  natural language inference.
\newblock In \emph{Proceedings of the 16th Conference of the European Chapter
  of the Association for Computational Linguistics: Main Volume}, pages
  255--269.

\bibitem[{Schuster et~al.(2022)Schuster, Fisch, Gupta, Dehghani, Bahri, Tran,
  Tay, and Metzler}]{schuster2022confident}
Tal Schuster, Adam Fisch, Jai Gupta, Mostafa Dehghani, Dara Bahri, Vinh~Q Tran,
  Yi~Tay, and Donald Metzler. 2022.
\newblock Confident adaptive language modeling.
\newblock \emph{arXiv preprint arXiv:2207.07061}.

\bibitem[{Schuster et~al.(2021)Schuster, Fisch, Jaakkola, and
  Barzilay}]{schuster2021consistent}
Tal Schuster, Adam Fisch, Tommi Jaakkola, and Regina Barzilay. 2021.
\newblock Consistent accelerated inference via confident adaptive transformers.
\newblock \emph{arXiv preprint arXiv:2104.08803}.

\bibitem[{Shih et~al.(2023)Shih, Sadigh, and Ermon}]{shih2023long}
Andy Shih, Dorsa Sadigh, and Stefano Ermon. 2023.
\newblock Long horizon temperature scaling.
\newblock \emph{arXiv preprint arXiv:2302.03686}.

\bibitem[{Si et~al.(2023)Si, Goyal, Wu, Zhao, Feng, Daum{\'e}~III, and
  Boyd-Graber}]{si2023large}
Chenglei Si, Navita Goyal, Sherry~Tongshuang Wu, Chen Zhao, Shi Feng, Hal
  Daum{\'e}~III, and Jordan Boyd-Graber. 2023.
\newblock Large language models help humans verify truthfulness--except when
  they are convincingly wrong.
\newblock \emph{arXiv preprint arXiv:2310.12558}.

\bibitem[{Socher et~al.(2013)Socher, Perelygin, Wu, Chuang, Manning, Ng, and
  Potts}]{socher2013recursive}
Richard Socher, Alex Perelygin, Jean Wu, Jason Chuang, Christopher~D Manning,
  Andrew~Y Ng, and Christopher Potts. 2013.
\newblock Recursive deep models for semantic compositionality over a sentiment
  treebank.
\newblock In \emph{Proceedings of the 2013 conference on empirical methods in
  natural language processing}, pages 1631--1642.

\bibitem[{Talmor et~al.(2018)Talmor, Herzig, Lourie, and
  Berant}]{talmor2018commonsenseqa}
Alon Talmor, Jonathan Herzig, Nicholas Lourie, and Jonathan Berant. 2018.
\newblock Commonsenseqa: A question answering challenge targeting commonsense
  knowledge.
\newblock \emph{arXiv preprint arXiv:1811.00937}.

\bibitem[{Tian et~al.(2023)Tian, Mitchell, Zhou, Sharma, Rafailov, Yao, Finn,
  and Manning}]{tian2023just}
Katherine Tian, Eric Mitchell, Allan Zhou, Archit Sharma, Rafael Rafailov,
  Huaxiu Yao, Chelsea Finn, and Christopher~D Manning. 2023.
\newblock Just ask for calibration: Strategies for eliciting calibrated
  confidence scores from language models fine-tuned with human feedback.
\newblock \emph{arXiv preprint arXiv:2305.14975}.

\bibitem[{Touvron et~al.(2023{\natexlab{a}})Touvron, Lavril, Izacard, Martinet,
  Lachaux, Lacroix, Rozi{\`e}re, Goyal, Hambro, Azhar
  et~al.}]{touvron2023llama}
Hugo Touvron, Thibaut Lavril, Gautier Izacard, Xavier Martinet, Marie-Anne
  Lachaux, Timoth{\'e}e Lacroix, Baptiste Rozi{\`e}re, Naman Goyal, Eric
  Hambro, Faisal Azhar, et~al. 2023{\natexlab{a}}.
\newblock Llama: Open and efficient foundation language models.
\newblock \emph{arXiv preprint arXiv:2302.13971}.

\bibitem[{Touvron et~al.(2023{\natexlab{b}})Touvron, Martin, Stone, Albert,
  Almahairi, Babaei, Bashlykov, Batra, Bhargava, Bhosale
  et~al.}]{touvron2023llama2}
Hugo Touvron, Louis Martin, Kevin Stone, Peter Albert, Amjad Almahairi, Yasmine
  Babaei, Nikolay Bashlykov, Soumya Batra, Prajjwal Bhargava, Shruti Bhosale,
  et~al. 2023{\natexlab{b}}.
\newblock Llama 2: Open foundation and fine-tuned chat models.
\newblock \emph{arXiv preprint arXiv:2307.09288}.

\bibitem[{Von~Oswald et~al.(2023)Von~Oswald, Niklasson, Randazzo, Sacramento,
  Mordvintsev, Zhmoginov, and Vladymyrov}]{von2023transformers}
Johannes Von~Oswald, Eyvind Niklasson, Ettore Randazzo, Jo{\~a}o Sacramento,
  Alexander Mordvintsev, Andrey Zhmoginov, and Max Vladymyrov. 2023.
\newblock Transformers learn in-context by gradient descent.
\newblock In \emph{International Conference on Machine Learning}, pages
  35151--35174. PMLR.

\bibitem[{Voorhees and Tice(2000)}]{voorhees2000building}
Ellen~M Voorhees and Dawn~M Tice. 2000.
\newblock Building a question answering test collection.
\newblock In \emph{Proceedings of the 23rd annual international ACM SIGIR
  conference on Research and development in information retrieval}, pages
  200--207.

\bibitem[{Wei et~al.(2022)Wei, Wang, Schuurmans, Bosma, Chi, Le, and
  Zhou}]{wei2022chain}
Jason Wei, Xuezhi Wang, Dale Schuurmans, Maarten Bosma, Ed~Chi, Quoc Le, and
  Denny Zhou. 2022.
\newblock Chain of thought prompting elicits reasoning in large language
  models.
\newblock \emph{arXiv preprint arXiv:2201.11903}.

\bibitem[{Xie et~al.(2021)Xie, Raghunathan, Liang, and Ma}]{xie2021explanation}
Sang~Michael Xie, Aditi Raghunathan, Percy Liang, and Tengyu Ma. 2021.
\newblock An explanation of in-context learning as implicit bayesian inference.
\newblock \emph{arXiv preprint arXiv:2111.02080}.

\bibitem[{Zhang et~al.(2015)Zhang, Zhao, and LeCun}]{zhang2015character}
Xiang Zhang, Junbo Zhao, and Yann LeCun. 2015.
\newblock Character-level convolutional networks for text classification.
\newblock \emph{Advances in neural information processing systems}, 28.

\bibitem[{Zhang et~al.(2020)Zhang, Liao, and Bellamy}]{zhang2020effect}
Yunfeng Zhang, Q~Vera Liao, and Rachel~KE Bellamy. 2020.
\newblock Effect of confidence and explanation on accuracy and trust
  calibration in ai-assisted decision making.
\newblock In \emph{Proceedings of the 2020 conference on fairness,
  accountability, and transparency}, pages 295--305.

\bibitem[{Zhao et~al.(2023)Zhao, Wei, Preston, and Poon}]{zhao2023automatic}
Theodore Zhao, Mu~Wei, J~Samuel Preston, and Hoifung Poon. 2023.
\newblock Automatic calibration and error correction for large language models
  via pareto optimal self-supervision.
\newblock \emph{arXiv preprint arXiv:2306.16564}.

\bibitem[{Zhao et~al.(2021)Zhao, Wallace, Feng, Klein, and
  Singh}]{zhao2021calibrate}
Zihao Zhao, Eric Wallace, Shi Feng, Dan Klein, and Sameer Singh. 2021.
\newblock Calibrate before use: Improving few-shot performance of language
  models.
\newblock In \emph{ICML}.

\bibitem[{Zheng et~al.(2023)Zheng, Chiang, Sheng, Zhuang, Wu, Zhuang, Lin, Li,
  Li, Xing, Zhang, Gonzalez, and Stoica}]{zheng2023judging}
Lianmin Zheng, Wei-Lin Chiang, Ying Sheng, Siyuan Zhuang, Zhanghao Wu, Yonghao
  Zhuang, Zi~Lin, Zhuohan Li, Dacheng Li, Eric.~P Xing, Hao Zhang, Joseph~E.
  Gonzalez, and Ion Stoica. 2023.
\newblock \href {http://arxiv.org/abs/2306.05685} {Judging llm-as-a-judge with
  mt-bench and chatbot arena}.

\bibitem[{Zhou et~al.(2023{\natexlab{a}})Zhou, Wan, Proleev, Mincu, Chen,
  Heller, and Roy}]{zhou2023batch}
Han Zhou, Xingchen Wan, Lev Proleev, Diana Mincu, Jilin Chen, Katherine Heller,
  and Subhrajit Roy. 2023{\natexlab{a}}.
\newblock Batch calibration: Rethinking calibration for in-context learning and
  prompt engineering.
\newblock \emph{arXiv preprint arXiv:2309.17249}.

\bibitem[{Zhou et~al.(2023{\natexlab{b}})Zhou, Jurafsky, and
  Hashimoto}]{zhou2023navigating}
Kaitlyn Zhou, Dan Jurafsky, and Tatsunori Hashimoto. 2023{\natexlab{b}}.
\newblock Navigating the grey area: Expressions of overconfidence and
  uncertainty in language models.
\newblock \emph{arXiv preprint arXiv:2302.13439}.

\bibitem[{Ziegler et~al.(2019)Ziegler, Stiennon, Wu, Brown, Radford, Amodei,
  Christiano, and Irving}]{ziegler2019fine}
Daniel~M Ziegler, Nisan Stiennon, Jeffrey Wu, Tom~B Brown, Alec Radford, Dario
  Amodei, Paul Christiano, and Geoffrey Irving. 2019.
\newblock Fine-tuning language models from human preferences.
\newblock \emph{arXiv preprint arXiv:1909.08593}.

\end{thebibliography}
\bibliographystyle{acl_natbib}

\appendix
\newpage
\onecolumn

\section{Extended Related Work}
\textbf{Uncertainty quantification in NLP.} 
Uncertainty quantification in NLP, which often adopts the Bayesian principle to sophisticated methods tailored for neural networks, aims to enhance the reliability of model predictions. This may involve non-trivial designs as directly interpreting language model predictions via probabilities \citep{kadavath2022language} and linguistic expressions \citep{lin2022teaching, mielke2022reducing, zhou2023navigating} may inadvertently lead to over-reliance on the model's uncertainties \citep{si2023large}, thus complicating the establishment of trustworthy common ground between humans and models \citep{buccinca2021trust}.
Notable recent advancements include employing model confidence as a critical factor in various applications like dialogue generation \citep{mielke2022reducing}, cascading prediction \citep{schuster2021consistent}, open-domain QA \citep{fisch2020efficient, angelopoulos2022conformal}, summarization \citep{laban2022summac}, language modeling \citep{schuster2022confident}, image captioning \citep{petryk2023simple}. 
\vspace{-0.1cm}
\section{Additional Experimental Details}
\begin{figure*}[thb]
\vspace{-0.3cm}
\begin{minipage}[t]{\textwidth}
\centering
\subfigure{
\includegraphics[width=.31\textwidth]{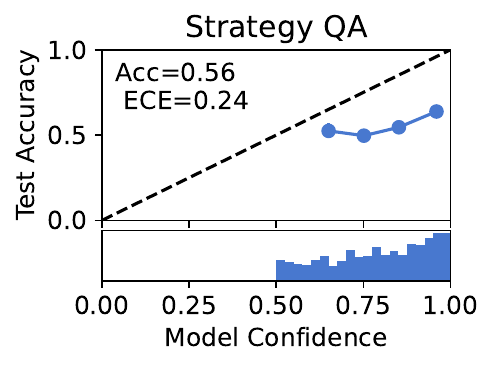}
}
\subfigure{
\includegraphics[width=.31\textwidth]{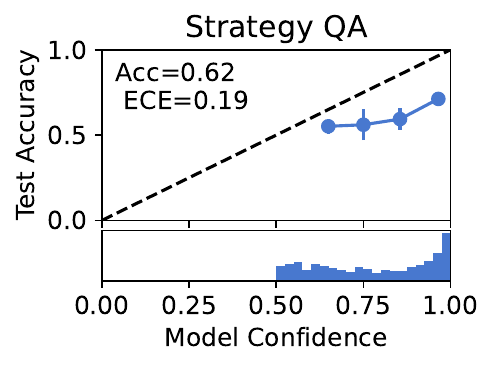} 
}
\subfigure{
\includegraphics[width=.31\textwidth]{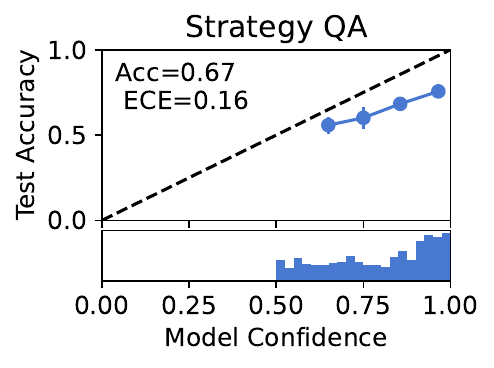} 
}
\end{minipage}
\vspace{-0.3cm}
\begin{minipage}[t]{\textwidth}
\centering
\subfigure{
\includegraphics[width=.31\textwidth]{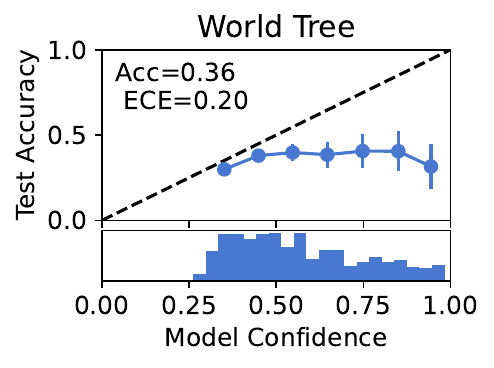}
}
\subfigure{
\includegraphics[width=.31\textwidth]{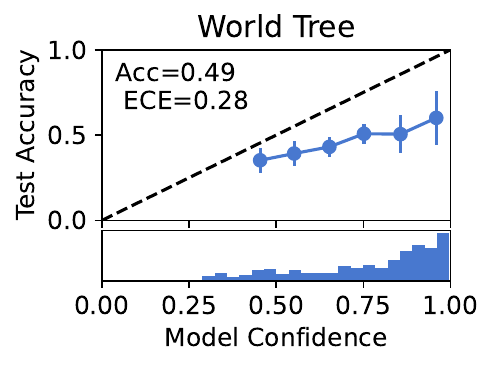} 
}
\subfigure{
\includegraphics[width=.31\textwidth]{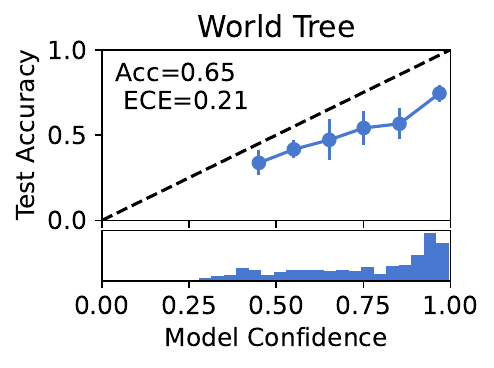} 
}
\end{minipage}
\caption{
Reliability plots and confidence histograms of LLaMA models on 4-shot reasoning tasks. 
Results of different sizes 7B (left), 13B (middle), and 30B (right) are plotted.
}
\vspace{-0.3cm}
\label{fig:reliability_appendix}
\end{figure*}

We provide prompts we adopt for experiments in Tab.\ref{tab:question_classification}. Additional reliability plots are shown in Fig.~\ref{fig:reliability_appendix}. 
Moreover, we provide extra results that extend those in the main text. Our implementation is open-sourced at \url{https://github.com/hlzhang109/icl-calibration}. 
The greatest accuracy and ECE values are highlighted in \textbf{bold} and \resred{red}, respectively. Extremely poor performance due to length truncation is omitted.

\noindent
\textbf{Model performance and calibration.} We present experimental results considering different model sizes for text classification and reasoning in Tables \ref{tab:main_cls} and \ref{tab:main_cot}, respectively. 
With the increase in model sizes, we observed overall improvements in model performance across most datasets. 
However, the calibration error (ECE) did not decrease immediately: for low-shot settings where $k<4$, models tend to have an ECE larger than $0.1$. 
On the other hand, ECE can decrease given more ICL examples ($k=8$) if context length is adequate.
Overall, zero-shot ICL can lead to good calibration results though the predictive performance is substantially weaker.
Interestingly, for some benchmarks like SST-2 and OpenBook QA, the ECE of the 30B model even surpassed that of the 7B model. 
Moreover, the ECE curves of the 7B and 13B models exhibited similar patterns to the 30B results as the number of ICL samples increased, as shown in the main Tab. (\ref{tab:main_tab}). 

\noindent
\textbf{The effect of fine-tuning.} We provide full results of all finetuned LLMs in \cref{tab:mean_std_baselines}, complementing Fig. (\ref{fig:finetuned}). 
We reach a similar conclusion as we explain in the main text: with an increasing number of ICL examples, accuracy generally improves but ECE first increases then decreases and miscalibration is widespread; 
an MoE model can also have the same accuracy-calibration trade-off;
fine-tuning substantially improves accuracy but hurts calibration by a large margin.

\noindent
\textbf{Results reliability.} Furthermore, as prompting is susceptible to various forms of biases and noises \citep{zhao2021calibrate, han2023prototypical, fei2023mitigating, zhou2023batch}, to provide a comprehensive understanding of the experimental outcomes, we delve into the variance across all experimental repetitions. 
Table \ref{tab:std_main} provides a detailed analysis of the variance metrics, affirming the stability and reliability of our experimental findings.

\begin{algorithm}
  \caption{Pseudocode for temperature scaling}
  \label{alg:temperature_scaling}
  \KwData{%
    $\mathcal{P}_\theta(\textbf{w})$: Original output of the classification model, $\mathcal{D}$: Training dataset, $\tau$: Temperature parameter, $k$: we use $k$-shot experimental settings, where during test the ICL prompts will consist of $k$ (sample, label) pairs.
  }

  \KwResult{%
    Adjusted probabilities after temperature scaling\;
  }

  \BlankLine

  \tcp{Training process}
  \tcp{$0$-shot: $(\mathbf{w}_i, y_i)$ is every training samples and corresponding label. }
  \tcp{$k$-shot: $\mathbf{w}_i=\{x_1,y_1,...,x_k,y_k, x_i\}$ uses $k$ prompt pairs.}
  \tcp{Fix $w$: the prompt in $\mathbf{w}_i=\{x_1,y_1,...,x_k,y_k, x_i\}$ will be used for all training instances and used during inference.}
  \For{each training sample $(\mathbf{w}_i, y_i) \in \mathcal{D}$}{
    Compute the original output: $z_i = \mathcal{P}_\theta(\mathbf{w}_i; \theta)$\;
    Compute the cross-entropy loss: $L_i = \text{CrossEntropy}(z_i, y_i)$\;
  }

  Compute the gradient of the loss to the temperature parameter: $\nabla_\tau \mathcal{L} = \frac{1}{|\mathcal{D}|} \sum_{i=1}^{|\mathcal{D}|} \nabla_\tau L_i$\;
  Update the temperature parameter using gradient descent: $\tau \leftarrow \tau - \eta \nabla_\tau \mathcal{L}$\;

  \BlankLine

  \tcp{Test time}
  \For{each test sample $\mathbf{x}_j$}{
    Compute the original output with prompt: $z_j = \mathcal{P}_\theta(\mathbf{x}_j; \theta)$\;
    Compute the adjusted output: $\hat{z}_j = \frac{z_j}{\tau}$\;
    Compute the softmax probabilities: $\hat{p}_j = \text{Softmax}(\hat{z}_j)$\;
  }
\end{algorithm}

\begin{table*}[h]
\centering
\resizebox{\linewidth}{!}{%
\begin{tabular}{lcc}
\toprule
\textbf{Dataset} & \textbf{Prompt} & \textbf{Label}\\ \cmidrule{1-3}
SST-2 &  \multicolumn{1}{|m{16cm}|}{Review: it may not be a great piece of filmmaking, but its power comes from its soul's - eye view of how well-meaning patronizing masked a social injustice, at least as represented by this case .\newline Sentiment: Positive \newline \newline Review: smith's point is simple and obvious -- people's homes are extensions of themselves, and particularly eccentric people have particularly eccentric living spaces -- but his subjects are charmers .\newline Sentiment:} & Negative, Positive\\
\hline
CB &  \multicolumn{1}{|m{16cm}|}{A: No, not really. I spend a lot of time with our income tax, though. especially, this year and last year. Um, I have been married for just a few years, so I've had to really switch around from the EZ form to the, uh, B: Schedule A. A: Right. B: Well, yeah. A: All the deductions and all that. B: Did you notice that when they passed the new simplified tax act, it seemed like it made everything harder?\newline question: when they passed the new simplified tax act it seemed like it made everything harder. true, false, or neither?\newline answer: true\newline \newline There was a group of curious onlookers... Marie felt her legs give way beneath her, she sat down on the edge of the pavement, feet in the gutter, doubled-up, sick and winded as if someone had punched her in the stomach. She lifted up her head and looked again. She had watched scenes like this so often in detective films and police series on television that she could hardly believe that this was real life.\newline question: this was real life. true, false, or neither?\newline answer:} &  True, False, Neither \\ \hline
RTE & \multicolumn{1}{|m{16cm}|}{ The main institutionalised forms of recognition for those who have made a significant contribution in the fields of physics, chemistry, medicine, literature, as well as for those working for peace (and more recently in the area of economics), are the Nobel prizes.\newline question: Nobel Peace Prize candidates have been chosen. True or False?\newline answer: False\newline\newline Egypt on Thursday strongly criticized Israeli new Foreign Minister Avigdor Lieberman for his remarks that he refused to recognize the peace efforts initiated in 2007 in the U.S. city of Annapolis to restore the peace talks with the Palestinians, reported the state MENA news agency. Lieberman\'s remarks is "regrettable," Egyptian Foreign Ministry spokesman Hossam Zaki was quoted as saying, adding "his remarks are the first blow to the peace efforts to come from the Israeli new government."\newline question: Hossam Zaki is the new Foreign Minister of Israel. True or False?\newline answer:} &  True, False \\ \hline
Strategy QA &  \multicolumn{1}{|m{16cm}|}{Question: Can spiders help eggplant farmers control parasites? Choose the answer from True and False.\newline Answer: The potato tuber moth is a parasite that targets the plant family Solanaceae, including eggplant Selenops radiatus is a spider genus in South Africa that effectively controls the potato tuber moth So, the answer is: True \newline \newline Question: Is the voice of Genie from Disney's Aladdin still alive? Choose the answer from True and False.\newline Answer:} &  True, False \\ \hline
Commonsense QA & \multicolumn{1}{|m{16cm}|}{"Question: Dan was a farmer with just one heifer.  But that was okay, he only kept her for milk, and he didn't think he'd find good farmland in a place as cold as where?\newline A arizona\newline  B farm yard\newline C michigan\newline D german field\newline E dairy farm\newline  Answer: Michigan is a state in the us where it precipitates throughout the year and areas, where it precipitates throughout the year, are generally cold. So the farmer thought he'd not find a good farmland in a place as cold as michigan. Enslaving heifers or other animals for their milk is wrong as they want to live free. All the places in the other options may not be cold. So, the answer is: C\newline \newline Question: From where does a snowflake form?\newline A cloud\newline B snow storm\newline C billow\newline D air\newline E snowstorm\newline Answer:"} &  A, B, C, D, E \\ \hline
\end{tabular}
}
\caption{\textbf{Prompts used for text classification and reasoning tasks}, with a single training example showcased per task for illustrative purposes. 
The right column displays corresponding labels. 
The prompting formats and labels for WorldTree and OpenBookQA are the same as those of the CommonsenseQA dataset.
}
\label{tab:question_classification}
\end{table*}

\begin{table}[]
\centering
\begin{tabular}{ccccccccc}
\toprule
\hline
{Metric} & {Dataset} & {Model Size} & 0-shot & 1-shot & 2-shot & 3-shot & 4-shot & 8-shot \\ \hline
\multirow{12}{*}{{ECE}} & \multirow{3}{*}{{AGNews}}  &  {7B} & 0.067 & 0.105 & \resred{0.225} & 0.158 & 0.086 & 0.075 \\ 
 & & {13B} & 0.093 & 0.084 & 0.069 & \resred{0.121} & 0.103 & 0.045 \\
  &  & {30B} & 0.261 & 0.043 & 0.049 & \resred{0.067} & 0.049 & 0.047 \\
 \cmidrule{3-9}
 & \multirow{3}{*}{{CB}}  & {7B} & 0.133 & \resred{0.218} & 0.172 & 0.197 & 0.202 & 0.215 \\
 &   & {13B} & 0.029 & 0.257 & \resred{0.282} & 0.221 & 0.263 & 0.216 \\ 
  &  & {30B} & 0.069 & \resred{0.312} & 0.216 & 0.217 & 0.192 & 0.181 \\
\cmidrule{3-9}
 & \multirow{3}{*}{{RTE}}  & {7B} & 0.068 & 0.075 & \resred{0.098} & 0.112 & 0.091 & 0.064 \\
 &  & {13B} & 0.042 & \resred{0.104} & 0.048 & 0.048 & 0.049 & 0.050 \\
 &  & {30B} & 0.023 & 0.051 & \resred{0.060} & 0.050 & 0.048 & 0.058 \\
 \cmidrule{3-9}
 & \multirow{3}{*}{{SST-2}} & {7B} & 0.038 & \resred{0.142} & 0.132 & 0.121 & 0.108 & 0.064 \\ 
 &  & {13B} & 0.051 & \resred{0.134} & 0.108 & 0.084 & 0.073 & 0.053 \\ 
 &  & {30B} & 0.083 & \resred{0.163} & 0.139 & 0.126 & 0.112 & 0.080 \\\midrule
\multirow{12}{*}{{ACC}} & \multirow{3}{*}{{AGNews}} & {7B} & 0.447 & 0.629 & 0.563 & 0.630 & 0.777 & \textbf{0.833} \\
 &   & {13B} & 0.490 & 0.812 & 0.773 & 0.720 & 0.775 & \textbf{0.847} \\ 
  &  & {30B} & 0.370 & 0.830 & 0.817 & 0.810 & 0.821 & \textbf{0.855} \\
\cmidrule{3-9}
 & \multirow{3}{*}{{CB}}& {7B} & 0.482 & 0.596 & 0.675 & 0.696 & 0.691 & \textbf{0.729} \\ 
 &  & {13B} & 0.554 & 0.627 & 0.659 & 0.691 & 0.611 & \textbf{0.709} \\  
  &  & {30B} & 0.500 & 0.696 & 0.789 & \textbf{0.834} & 0.814 & 0.796 \\
\cmidrule{3-9}
 & \multirow{3}{*}{{RTE}} & {7B} & 0.552 & 0.668 & 0.653 & 0.646 & 0.653 & \textbf{0.698} \\ 
 &   & {13B} & 0.679 & 0.673 & 0.708 & 0.723 & 0.723 & \textbf{0.746} \\ 
  &  & {30B} & 0.672 & 0.742 & 0.747 & 0.738 & 0.748 & \textbf{0.752} \\
\cmidrule{3-9}
 & \multirow{3}{*}{{SST-2}} & {7B} & 0.483 & 0.799 & 0.877 & 0.908 & 0.917 & \textbf{0.954} \\  
 & & {13B} & 0.483 & 0.918 & 0.943 & 0.955 & 0.962 & \textbf{0.969} \\
  &  & {30B} & 0.607 & 0.930 & 0.940 & 0.961 & \textbf{0.964} & \textbf{0.964} \\
\hline
 \bottomrule
\end{tabular}%
\caption{{Accuracy and Calibration} of LLaMA model with three sizes across four text classification datasets.}
\label{tab:main_cls}
\end{table}

\begin{table}[]
\centering
\begin{tabular}{ccccccccc}
\toprule
\hline
{Metric} & {Dataset} & {Model Size} & 0-shot & 1-shot & 2-shot & 3-shot & 4-shot & 8-shot \\ \hline
\multirow{12}{*}{{ECE}} & \multirow{3}{*}{{Commonsense QA}} &  {7B} & 0.070 & 0.155 & 0.237 & 0.227 & 0.238 & - \\
 & & {13B} & 0.066 & 0.161 & 0.282 & 0.292 & 0.310 & - \\
  &  & {30B} & 0.048 & 0.232 & \resred{0.290} & 0.253 & 0.283 & - \\ \cmidrule{3-9}
 & \multirow{3}{*}{{OpenBook QA}} & {7B} & 0.040 & 0.241 & \resred{0.270} & 0.184 & 0.130 & 0.121 \\
  & & {13B} & 0.031 & 0.132 & \resred{0.217} & 0.209 & 0.191 & 0.175 \\
  &  & {30B} & 0.048 & 0.232 & \resred{0.290} & 0.253 & 0.283 & - \\
  \cmidrule{3-9}
 & \multirow{3}{*}{{Strategy QA}}  &   {7B} & 0.133 & \resred{0.275} & 0.206 & 0.243 & 0.242 & 0.227 \\ 
 & & {13B} & 0.051 & 0.154 & 0.170 & \resred{0.192} & 0.188 & 0.190 \\
 &  & {30B} & 0.204 & 0.154 & 0.174 & 0.172 & 0.161 & \resred{0.193} \\
 \cmidrule{3-9}
 & \multirow{3}{*}{{World Tree}} & {13B} & 0.065 & 0.113 & 0.226 & 0.250 & 0.284 & - \\
 &  & {30B} & 0.112 & 0.211 & 0.251 & 0.185 & 0.206 & - \\
 &  & {7B} & 0.074 & 0.124 & 0.198 & 0.179 & 0.203 & - \\ \hline
\multirow{12}{*}{{ACC}} & \multirow{3}{*}{{Commonsense QA}}  &   {7B} & 0.224 & 0.292 & 0.388 & \textbf{0.421} & 0.406 & - \\ 
&  & {13B} & 0.320 & 0.478 & 0.549 & \textbf{0.574} & 0.562 & - \\
 &  & {30B} & 0.356 & 0.589 & 0.608 & \textbf{0.675} & 0.644 & - \\
\cmidrule{3-9}
 & \multirow{3}{*}{{OpenBook QA}}  & {7B} & 0.308 & 0.298 & 0.376 & 0.417 & 0.454 & \textbf{0.480} \\
 & & {13B} & 0.362 & 0.454 & 0.509 & 0.551 & 0.580 & \textbf{0.611} \\
  &  & {30B} & 0.386 & 0.561 & 0.604 & 0.644 & 0.648 & \textbf{0.662} \\
\cmidrule{3-9}
 & \multirow{3}{*}{{Strategy QA}} & {7B} & 0.566 & 0.488 & 0.554 & 0.550 & 0.562 & \textbf{0.575} \\ 
 &  & {13B} & 0.554 & 0.598 & \textbf{0.621} & 0.595 & 0.618 & 0.612 \\
  &  & {30B} & 0.450 & 0.619 & 0.654 & 0.660 & \textbf{0.672} & 0.662 \\
\cmidrule{3-9}
 & \multirow{3}{*}{{World Tree}}  & {7B} & 0.302 & 0.298 & 0.326 & \textbf{0.384} & 0.362 & - \\ 
  & & {13B} & 0.444 & 0.437 & 0.495 & \textbf{0.519} & 0.492 & - \\
   &  & {30B} & 0.534 & 0.570 & 0.621 & \textbf{0.680} & 0.646 & - \\
 \hline
 \bottomrule
\end{tabular}%
\caption{{Accuracy and Calibration} of LLaMA models across three sizes on four reasoning datasets.}
\label{tab:main_cot}
\end{table}

\begin{table}[]
\resizebox{\textwidth}{!}{%
\begin{tabular}{cccccccc}
\toprule
\hline
{Dataset} & {Metric} & {0-shot} & {1-shot} & {2-shot} & {3-shot} & {4-shot} & {8-shot} \\ \hline
\multirow{2}{*}{CB} & ACC & $0.500_{\pm 0.000}$ & $0.696_{\pm0.304}$ & $0.789_{\pm0.138}$ & $\mathbf{0.834_{\pm0.068}}$ & $0.814_{\pm0.068}$ & $0.796_{\pm 0.110}$ \\
 & ECE & $0.143_{\pm 0.000}$ & $\resred{0.409_{\pm 0.041}}$ & $0.216_{\pm 0.061}$ & $0.217_{\pm0.057}$ & $0.376_{\pm 0.053}$ & $0.359_{\pm 0.071}$ \\\hline 
\multirow{2}{*}{RTE} & ACC & $0.672_{\pm0.000}$ & $0.742_{\pm0.018}$ & $0.747_{\pm0.032}$ & $0.738_{\pm0.044}$ & $0.748_{\pm0.043}$ & $\mathbf{0.752_{\pm0.039}}$ \\
 & ECE & $0.023_{\pm0.000}$ & $0.051_{\pm0.020}$ & $\resred{0.060_{\pm 0.021}}$ & $0.050_{\pm 0.023}$ & $0.048_{\pm 0.017}$ & $0.058_{\pm 0.022}$ \\\hline 
\multirow{2}{*}{SST-2} & ACC & $0.607_{\pm 0.000}$ & $0.930_{\pm 0.025}$ & $0.940_{\pm 0.066}$ & $0.961_{\pm0.017}$ & $\mathbf{0.964_{\pm 0.012}}$ & $\mathbf{0.964_{\pm0.011}}$ \\
 & ECE & $0.106_{\pm0.000}$ & $\resred{0.339_{\pm0.026}}$ & $0.139_{\pm 0.058}$ & $0.126_{\pm 0.053}$ & $0.310_{\pm 0.022}$ & $0.287_{\pm 0.014}$ \\ \hline 
\multirow{2}{*}{AGnews} & ACC & $0.370_{\pm 0.000}$ & $0.830_{\pm 0.015}$ & $0.817_{\pm 0.017}$ & $0.810_{\pm 0.056}$ & $0.821_{\pm 0.029}$ & $\mathbf{0.855_{\pm 0.017}}$ \\
 & ECE & $0.261_{\pm 0.000}$ & $0.043_{\pm 0.009}$ & $0.049_{\pm0.016}$ & $\resred{0.067_{\pm0.029}}$ &  $0.049_{\pm 0.017}$ & $0.047_{\pm0.018}$ \\ \hline
\multirow{2}{*}{OpenBook QA} & ACC & $0.386_{\pm 0.000}$ & $0.561_{\pm 0.028}$ & $0.604_{\pm 0.027}$ & $0.644_{\pm 0.016}$ & $0.648_{\pm 0.018}$ & $\mathbf{0.662_{\pm 0.031}}$ \\
 & ECE & $0.036_{\pm 0.000}$ & $0.231_{\pm 0.049}$ & $\resred{0.255_{\pm 0.050}}$ & $0.207_{\pm 0.041}$ & $0.206_{\pm 0.019}$ & $0.191_{\pm 0.022}$ \\\hline 
\multirow{2}{*}{CommonSense QA} & ACC & $0.356_{\pm 0.000}$ & $0.586_{\pm 0.028}$ & $0.608_{\pm 0.013}$ & $\mathbf{0.675_{\pm 0.027}}$ & $0.644_{\pm 0.034}$ & $0.653_{\pm 0.090}$ \\
 & ECE & $0.048_{\pm 0.000}$ & $0.232_{\pm 0.102}$ & $\resred{0.290_{\pm 0.022}}$ & $0.253_{\pm 0.028}$ & $0.283_{\pm 0.045}$ & $0.289_{\pm 0.140}$ \\\hline 
\multirow{2}{*}{Strategy QA} & ACC & $0.450_{\pm 0.000}$ & $0.619_{\pm 0.030}$ & $0.654_{\pm 0.033}$ & $0.660_{\pm 0.022}$ & $\mathbf{0.672_{\pm 0.015}}$ & - \\
 & ECE & $0.204_{\pm 0.000}$ & $0.154_{\pm 0.029}$ & $\resred{0.174_{\pm 0.070}}$ & $0.172_{\pm 0.025}$ & $0.161_{\pm 0.008}$ & - \\\hline 
\multirow{2}{*}{World Tree} & ACC & $0.554_{\pm 0.000}$ & $0.570_{\pm 0.056}$ & $0.621_{\pm 0.109}$ & $\mathbf{0.680_{\pm 0.072}}$ & $0.504_{\pm 0.074}$ & - \\
 & ECE & $0.112_{\pm 0.000}$ & $0.211_{\pm 0.042}$ & $\resred{0.251_{\pm 0.101}}$ & $0.185_{\pm 0.048}$ & $0.144_{\pm 0.051}$ & - \\
 \hline
 \bottomrule
\end{tabular}
}
\caption{The full results (mean and standard deviation) for various experimental configurations extending Table.~\ref{tab:main_tab}.}
\label{tab:std_main}
\end{table}

\begin{table}[]
\resizebox{\textwidth}{!}{%
\begin{tabular}{cccccccc}
\toprule
\hline
{Dataset} & {Metric} & {0-shot} & {1-shot} & {2-shot} & {3-shot} & {4-shot} & {8-shot} \\ \hline
\multicolumn{8}{c}{CB} \\ 
\toprule
\multirow{2}{*}{Alpaca-7B} & ACC & $0.552_{\pm 0.000}$ & $0.668_{\pm 0.032}$ & $0.653_{\pm 0.079}$ & $0.646_{\pm0.086}$ & $0.653_{\pm 0.067}$ & $\mathbf{0.698_{\pm 0.028}}$ \\
 & ECE & $0.016_{\pm 0.000}$ & $0.119_{\pm 0.018}$ & $0.123_{\pm 0.044}$ & $0.122_{\pm 0.031}$ & $0.115_{\pm 0.017}$ & $\resred{0.127_{\pm 0.020}}$ \\
 \hline
\multirow{2}{*}{LLama2-Chat-7B} & ACC & $0.375_{\pm 0.000}$ & $0.566_{\pm 0.129}$ & $0.643_{\pm 0.107}$ & $0.670_{\pm 0.126}$ & $\mathbf{0.677_{\pm 0.113}}$ & $\mathbf{0.677_{\pm 0.111}}$ \\
 & ECE & $0.287_{\pm 0.000}$ & $\resred{0.223_{\pm 0.078}}$ & $0.170_{\pm 0.062}$ & $0.153_{\pm 0.054}$ & $0.154_{\pm 0.054}$ & $0.170_{\pm 0.054}$ \\
 \hline
\multirow{2}{*}{LLama2-7B} & ACC & $0.339_{\pm 0.000}$ & $0.464_{\pm 0.193}$ & $0.511_{\pm 0.163}$ & $0.538_{\pm 0.113}$ & $0.534_{\pm 0.109}$ & $\mathbf{0.575_{\pm 0.059}}$ \\
 & ECE & $0.125_{\pm 0.000}$ & $0.222_{\pm 0.190}$ & $0.174_{\pm 0.029}$ & $0.206_{\pm 0.066}$ & $\resred{0.226_{\pm 0.071}}$ & $0.222_{\pm 0.058}$ \\
 \hline
\multirow{2}{*}{Mistral-7B-v0.1} & ACC & $0.500_{\pm 0.000}$ & $0.643_{\pm 0.264}$ & $0.725_{\pm 0.198}$ & $\mathbf{0.827_{\pm 0.067}}$ & $0.793_{\pm 0.063}$ & $0.793_{\pm 0.121}$ \\
 & ECE & $0.063_{\pm 0.000}$ & $\resred{0.330_{\pm 0.118}}$ & $0.228_{\pm 0.094}$ & $0.244_{\pm 0.036}$ & $0.193_{\pm 0.048}$ & $0.144_{\pm 0.028}$ \\
 \hline
\multirow{2}{*}{vicuna-7b-v1.5} & ACC & $0.571_{\pm 0.000}$ & $0.668_{\pm 0.049}$ & $0.663_{\pm 0.052}$ & $0.668_{\pm 0.058}$ & $\mathbf{0.675_{\pm 0.061}}$ & $0.648_{\pm 0.073}$ \\
 & ECE & $0.051_{\pm 0.000}$ & $0.176_{\pm 0.034}$ & $0.172_{\pm 0.047}$ & $0.169_{\pm 0.054}$ & $0.170_{\pm 0.047}$ & $\resred{0.181_{\pm 0.052}}$ \\ \hline
\multicolumn{8}{c}{AGNews} \\
\toprule
\multirow{2}{*}{Alpaca-7B} & ACC & $0.810_{\pm 0.000}$ & $0.793_{\pm 0.041}$ & $0.710_{\pm 0.110}$ & $0.715_{\pm 0.111}$ & $0.782_{\pm 0.079}$ & $\mathbf{0.832_{\pm 0.029}}$ \\
& ECE & $0.043_{\pm 0.000}$ & $0.123_{\pm 0.033}$ & $\resred{0.190_{\pm 0.095}}$ & $0.167_{\pm 0.093}$ & $0.112_{\pm 0.057}$ & $0.065_{\pm 0.019}$ \\ \hline
\multirow{2}{*}{LLama2-Chat-7B} & ACC & $0.793_{\pm 0.000}$ & $0.809_{\pm 0.031}$ & $0.823_{\pm 0.046}$ & $0.829_{\pm 0.035}$ & $0.829_{\pm 0.028}$ & $\mathbf{0.843_{\pm 0.019}}$ \\
& ECE & $0.164_{\pm 0.000}$ & $\resred{0.162_{\pm 0.030}}$ & $0.143_{\pm 0.039}$ & $0.138_{\pm 0.033}$ & $0.138_{\pm 0.024}$ & $0.127_{\pm 0.013}$ \\ \hline
\multirow{2}{*}{LLama2-7B} & ACC & $0.573_{\pm 0.000}$ & $0.832_{\pm 0.022}$ & $0.789_{\pm 0.112}$ & $0.801_{\pm 0.108}$ & $0.849_{\pm 0.057}$ & $\mathbf{0.868_{\pm 0.009}}$ \\
& ECE & $0.102_{\pm 0.000}$ & $0.037_{\pm 0.012}$ & $0.074_{\pm 0.083}$ & $\resred{0.078_{\pm 0.082}}$ & $0.052_{\pm 0.024}$ & $0.053_{\pm 0.011}$ \\ \hline
\multirow{2}{*}{Mistral-7B-v0.1} & ACC & $0.780_{\pm 0.000}$ & $0.847_{\pm 0.017}$ & $0.842_{\pm 0.028}$ & $0.820_{\pm 0.056}$ & $0.808_{\pm 0.085}$ & $\mathbf{0.867_{\pm 0.004}}$ \\
& ECE & $0.193_{\pm 0.000}$ & $0.059_{\pm 0.012}$ & $0.044_{\pm 0.010}$ & $0.052_{\pm 0.022}$ & $\resred{0.077_{\pm 0.049}}$ & $0.043_{\pm 0.010}$ \\ \hline
\multirow{2}{*}{vicuna-7b-v1.5} & ACC & $0.740_{\pm 0.000}$ & $0.803_{\pm 0.013}$ & $0.834_{\pm 0.031}$ & $0.824_{\pm 0.054}$ & $\mathbf{0.835_{\pm 0.030}}$ & $0.832_{\pm 0.036}$ \\
& ECE & $0.063_{\pm 0.000}$ & $\resred{0.139_{\pm 0.012}}$ & $0.108_{\pm 0.025}$ & $0.116_{\pm 0.034}$ & $0.114_{\pm 0.014}$ & $0.109_{\pm 0.034}$ \\ \hline
\multicolumn{8}{c}{RTE} \\
\toprule
\multirow{2}{*}{Alpaca-7B} & ACC & $0.672_{\pm 0.000}$ & $0.644_{\pm 0.015}$ & $0.687_{\pm 0.019}$ & $0.696_{\pm 0.020}$ & $\mathbf{0.703_{\pm 0.015}}$ & $0.690_{\pm 0.035}$ \\
& ECE & $0.175_{\pm 0.000}$ & $\resred{0.270_{\pm 0.018}}$ & $0.212_{\pm 0.034}$ & $0.197_{\pm 0.028}$ & $0.184_{\pm 0.026}$ & $0.193_{\pm 0.025}$ \\ \hline
\multirow{2}{*}{LLama2-Chat-7B} & ACC & $0.729_{\pm 0.000}$ & $0.685_{\pm 0.042}$ & $0.687_{\pm 0.048}$ & $0.699_{\pm 0.040}$ & $0.709_{\pm 0.034}$ & $\mathbf{0.731_{\pm 0.033}}$ \\
& ECE & $0.165_{\pm 0.000}$ & $\resred{0.218_{\pm 0.031}}$ & $0.205_{\pm 0.033}$ & $0.198_{\pm 0.033}$ & $0.184_{\pm 0.030}$ & $0.172_{\pm 0.020}$ \\ \hline
\multirow{2}{*}{LLama2-7B} & ACC & $0.682_{\pm 0.000}$ & $0.684_{\pm 0.034}$ & $\mathbf{0.698_{\pm 0.049}}$ & $0.676_{\pm 0.058}$ & $0.689_{\pm 0.068}$ & $0.685_{\pm 0.050}$ \\
& ECE & $0.044_{\pm 0.000}$ & $0.076_{\pm 0.021}$ & $0.084_{\pm 0.029}$ & $0.085_{\pm 0.034}$ & $\resred{0.105_{\pm 0.031}}$ & $0.083_{\pm 0.032}$ \\ \hline
\multirow{2}{*}{Mistral-7B-v0.1} & ACC & $0.686_{\pm 0.000}$ & $0.731_{\pm 0.025}$ & $0.756_{\pm 0.015}$ & $0.768_{\pm 0.019}$ & $\mathbf{0.776_{\pm 0.016}}$ & $0.773_{\pm 0.025}$ \\
& ECE & $0.054_{\pm 0.000}$ & $\resred{0.121_{\pm 0.047}}$ & $0.080_{\pm 0.042}$ & $0.084_{\pm 0.033}$ & $0.087_{\pm 0.025}$ & $0.085_{\pm 0.035}$ \\ \hline
\multirow{2}{*}{vicuna-7b-v1.5} & ACC & $0.610_{\pm 0.000}$ & $0.731_{\pm 0.015}$ & $0.756_{\pm 0.011}$ & $0.762_{\pm 0.013}$ & $0.765_{\pm 0.019}$ & $\mathbf{0.770_{\pm 0.026}}$ \\
& ECE & $0.234_{\pm 0.000}$ & $\resred{0.101_{\pm 0.021}}$ & $0.073_{\pm 0.028}$ & $0.067_{\pm 0.016}$ & $0.057_{\pm 0.015}$ & $0.052_{\pm 0.015}$ \\ \hline
\multicolumn{8}{c}{SST-2} \\
\toprule
\multirow{2}{*}{Alpaca-7B} & ACC & $0.730_{\pm 0.000}$ & $0.868_{\pm 0.088}$ & $0.939_{\pm 0.018}$ & $0.949_{\pm 0.015}$ & $0.955_{\pm 0.012}$ & $\mathbf{0.952_{\pm 0.014}}$ \\
& ECE & $0.139_{\pm 0.000}$ & $\resred{0.068_{\pm 0.048}}$ & $0.025_{\pm 0.009}$ & $0.021_{\pm 0.006}$ & $0.020_{\pm 0.009}$ & $0.026_{\pm 0.010}$ \\ \hline
\multirow{2}{*}{LLama2-Chat-7B} & ACC & $0.867_{\pm 0.000}$ & $0.951_{\pm 0.008}$ & $0.942_{\pm 0.018}$ & $\mathbf{0.953_{\pm 0.012}}$ & $0.952_{\pm 0.016}$ & $0.952_{\pm 0.015}$ \\
& ECE & $0.039_{\pm 0.000}$ & $0.033_{\pm 0.006}$ & $\resred{0.044_{\pm 0.015}}$ & $0.032_{\pm 0.012}$ & $0.035_{\pm 0.014}$ & $0.037_{\pm 0.014}$ \\ \hline
\multirow{2}{*}{LLama2-7B} & ACC & $0.530_{\pm 0.000}$ & $0.754_{\pm 0.140}$ & $0.829_{\pm 0.121}$ & $0.874_{\pm 0.105}$ & $0.904_{\pm 0.062}$ & $\mathbf{0.925_{\pm 0.045}}$ \\
& ECE & $0.018_{\pm 0.000}$ & $0.180_{\pm 0.058}$ & $\resred{0.119_{\pm 0.076}}$ & $0.085_{\pm 0.072}$ & $0.062_{\pm 0.027}$ & $0.040_{\pm 0.012}$ \\ \hline
\multirow{2}{*}{Mistral-7B-v0.1} & ACC & $0.563_{\pm 0.000}$ & $0.958_{\pm 0.007}$ & $0.941_{\pm 0.058}$ & $0.956_{\pm 0.030}$ & $0.961_{\pm 0.022}$ & $\mathbf{0.969_{\pm 0.006}}$ \\
& ECE & $0.058_{\pm 0.000}$ & $0.133_{\pm 0.029}$ & $\resred{0.086_{\pm 0.033}}$ & $0.078_{\pm 0.032}$ & $0.072_{\pm 0.032}$ & $0.052_{\pm 0.020}$ \\ \hline
\multirow{2}{*}{vicuna-7b-v1.5} & ACC & $0.893_{\pm 0.000}$ & $0.946_{\pm 0.011}$ & $0.956_{\pm 0.008}$ & $0.961_{\pm 0.012}$ & $0.964_{\pm 0.008}$ & $\mathbf{0.968_{\pm 0.009}}$ \\
& ECE & $0.041_{\pm 0.000}$ & $\resred{0.032_{\pm 0.009}}$ & $0.025_{\pm 0.008}$ & $0.026_{\pm 0.006}$ & $0.023_{\pm 0.005}$ & $0.021_{\pm 0.005}$ \\ 
\hline
\bottomrule
\end{tabular}%
}
\caption{The full results (mean and standard deviation) for various experimental configurations extending Figure~\ref{fig:finetuned}.}
\label{tab:mean_std_baselines}
\end{table}

\begin{table*}[h]
    \resizebox{0.8\textwidth}{!}{%
    \begin{tabularx}{\textwidth}{|c|c|X|c|c|}
        \cline{1-5}
        Dataset & Case & \centering Reasoning & $\hat{y}$ & $p$ \\
        \cline{1-5}
        \multirow{12}{*}{Strategy QA} & \multicolumn{4}{c|}{Question: Is the voice of the Genie from Disney's Aladdin still alive? Choose the answer from True and False} \\
        \cline{2-5}
        & \multicolumn{4}{c|}{Options: A) True, B) False} \\
        \cline{2-5}
        & \multicolumn{4}{c|}{Ground Truth: A} \\
        \cline{2-5}
        & 7B & The voice of the Genie from Disney's Aladdin is still alive. & B & 0.6249 \\
        \cline{2-5}
        & 13B & The voice of the Genie from Disney's Aladdin is Robin Williams. Robin Williams died on August 11, 2014. & B & 0.7654 \\
        \cline{2-5}
        & 30B & Robin Williams was the voice of the Genie in Disney's Aladdin. He died on August 11, 2014. & B & 0.8471 \\
        \cline{1-5}
        \multirow{6}{*}{World Tree} & \multicolumn{4}{c|}{{\begin{tabular}[c]{@{}l@{}}Question: A bird eats a brightly-colored insect that tastes bad. \\ The bird avoids preying on this type of insect again. This example is best described as ?\end{tabular}}} \\
        \cline{2-5}
        & \multicolumn{4}{c|}{Options: A) a physical characteristic, B) a learned behavior, C) an inherited behavior, D) an adaptation} \\
        \cline{2-5}
        & \multicolumn{4}{c|}{Ground Truth: B} \\
        \cline{2-5}
        & 7B & {ing the juice from fruit is a kind of physical change. A physical change is a change in the physical state of a substance. A physical change is not a change in the chemical composition of a substance. A physical change is not a change in the chemical composition of a substance.} & D & 0.5249 \\
        \cline{2-5}
        & 13B & An adaptation is a change in an organism that makes it better suited to its environment.  & D & 0.7651 \\
        \cline{2-5}
        & 30B & Adaptation means a change in the body of an organism that helps it survive in its environment. Adaptation is a kind of behavior. & D & 0.8764 \\
        \cline{2-5} \hline
    \end{tabularx}
    }
    \caption{Qualitative Results of LLaMA on Strategy QA and WorldTree}
    \label{tab:confident_wrong_app}
\end{table*}

\end{document}